\documentclass[lettersize,journal]{IEEEtran}
\usepackage{amsmath,amsfonts}
\usepackage{algorithmic}
\usepackage{algorithm}
\usepackage{array}
\usepackage[caption=false,font=normalsize]{subfig}
\usepackage{textcomp}
\usepackage{stfloats}
\usepackage{url}
\usepackage{verbatim}
\usepackage{graphicx}
\usepackage{cite}
\usepackage{amsmath,amsthm, amssymb,mathtools}
\usepackage{orcidlink}
\usepackage{hyperref}
\usepackage{multirow}

\hypersetup{hidelinks=true}

\newtheorem{definition}{Definition}[section]
\newtheorem{theorem}{Theorem}[section]

\makeatletter
\newcommand*{\indep}{%
	\mathbin{%
		\mathpalette{\@indep}{}%
	}%
}
\newcommand*{\nindep}{%
	\mathbin{
		\mathpalette{\@indep}{\not}
	}%
}
\newcommand*{\@indep}[2]{%
	\sbox0{$#1\perp\m@th$}
	\sbox2{$#1=$}
	\sbox4{$#1\vcenter{}$}
	\rlap{\copy0}
	\dimen@=\dimexpr\ht2-\ht4-.2pt\relax
	\kern\dimen@
	{#2}%
	\kern\dimen@
	\copy0 
}

\begin{document}
\title{Causal Multi-Label Feature Selection \\ in Federated Setting}

\author{
    Yukun Song$^\dag$~\orcidlink{0009-0004-9843-3505}, 
    \and Dayuan Cao$^\dag$,
    \and Jiali Miao~\orcidlink{0009-0004-2718-4214}, 
    \and Shuai Yang~\orcidlink{0000-0002-1837-0515},
    \and Kui Yu~\orcidlink{0000-0003-2442-4572}
    \thanks{
        $\dag$ represents equal contribution.
    }
    \thanks{
        Yukun Song, Dayuan Cao, Jiali Miao and Kui Yu are with the School of Computer Science and Information Engineering, Hefei University of Technology, Hefei 230601, China (e-mail: \href{mailto:yukun_song@mail.hfut.edu.cn}{yukun\_song@mail.hfut.edu.cn}, \href{mailto:caodayuan@mail.hfut.edu.cn}{caodayuan@mail.hfut.edu.cn}, \href{mailto:miaojiali@mail.hfut.edu.cn}{miaojiali@mail.hfut.edu.cn}, \href{mailto:yukui@hfut.edu.cn}{yukui@hfut.edu.cn}).
    }
    \thanks{
        Shuai Yang is with the School of Information and Artificial Intelligence, Anhui Agricultural University, Hefei 230036, China (e-mail: \href{yangs@ahau.edu.cn}{yangs@ahau.edu.cn}).  
    }
}

\markboth{Journal of \LaTeX\ Class Files,~Vol.~00, No.~0, May~2024}%
{Shell \MakeLowercase{\textit{et al.}}: A Sample Article Using IEEEtran.cls for IEEE Journals}

\IEEEpubid{0000--0000/00\$00.00~\copyright~2024 IEEE}

\maketitle

\begin{abstract}
    Multi-label feature selection serves as an effective mean for dealing with high-dimensional multi-label data. To achieve satisfactory performance, existing methods for multi-label feature selection often require the centralization of substantial data from multiple sources. However, in federated setting, centralizing data from all sources and merging them into a single dataset is often infeasible. To tackle this issue, in this paper, we study a new problem of causal multi-label feature selection in federated setting and propose a Federated Causal Multi-label Feature Selection (FedCMFS) algorithm with three novel subroutines. Specifically, FedCMFS first uses the FedCFL subroutine that considers the correlations among label-label, label-feature, and feature-feature to learn the relevant features (candidate parents and children) of each class label while preserving data privacy without centralizing data. Second, FedCMFS employs the FedCFR subroutine to selectively recover the missed true relevant features. Finally, FedCMFS utilizes the FedCFC subroutine to remove false relevant features. The extensive experiments using eight datasets have validated the effectiveness of FedCMFS. 
\end{abstract}

\begin{IEEEkeywords}
     Multi-label data, Causal feature selection, Federated learning, Privacy preserving data, Parallel Optimization.
\end{IEEEkeywords}

\section{Introduction}\label{sec1}
    \IEEEPARstart{M}{ulti-label} learning has become an important research direction in the field of machine learning~~\cite{zhang2013review,tian2023causal}. As information technology rapidly develops, multi-label data is becoming increasingly complex, potentially leading to the problem of the curse of dimensionality~\cite{kashef2018multilabel}. Therefore, feature selection, as one of the effective tools to solve the curse of dimensionality, is widely used in multi-label learning~\cite{li2017feature}. It aims to reduce the dimensionality of features by designing a metric for feature importance, with the goal of selecting a subset of features that contains irrelevant or redundant features as few as possible.

    Existing multi-label feature selection algorithms are typically based on statistical co-occurrence relationships to determine feature dependency without providing an explanation for why they are dependent. To tackle this issue, researchers have proposed causal multi-label feature selection algorithms based on causal structures~\cite{wu2020multi}. The causal relationship describes the causal relationship between two variables, revealing the underlying mechanisms of how these variables interact. A causal structure always employ a Directed Acyclic Graphs (DAG) to represent causal relationships between variables. In a DAG, the existence of a directed edge from A to B represents that A is a parent (direct cause) of B, and conversely, B is a child (direct effect) of A~\cite{pearl2009causality}. Exploring the relationships between variables by learning the global causal structure can be computationally expensive, particularly when dealing with high-dimensional datasets. In contrast, local causal structure learning methods offer an efficient alternative. These methods directly identify the parents and children of a given label variable, making them more efficient compared to global causal structure algorithms.

    Currently those existing multi-label feature selection algorithms typically require access to all the data to determine important features and do not consider the data privacy. However, in many real-world application scenarios, data often originates from multiple sources, and the aggregation of data requires consideration of data privacy. For instance, chronic disease research may require patient data from various hospitals, leading to the risk of leaking patient data privacy. 
    
    To protect data privacy, federated learning has garnered considerable attention~\cite{mcmahan2017communication}. Federated learning builds machine learning models through multi-party collaboration and is primarily divided into vertical federated learning and horizontal federated learning. Vertical federated learning shares the same samples but each client holds different features, while clients in horizontal federated learning share the same feature space while holding different samples~\cite{yang2019federated}. In this study, we consider the horizontal federated learning setting. 

    Since currently there are no studies on multi-label feature selection for considering data privacy, to fill this gap, in this paper, we propose the Federated Causal Multi-label Feature Selection (FedCMFS) algorithm in federated setting, which comprises three subroutines: the Federated Causal Feature Learning  subroutine (FedCFL), the Federated Causal Feature Retrieval subroutine (FedCFR), and the Federated Causal Feature Correction subroutine (FedCFC). And we demonstrate the effectiveness of FedCMFS by simulating the federated setting with a large number of real datasets as well as conducting experiments using 5 comparison algorithms. 

\section{Related Work}\label{sec2}\IEEEpubidadjcol 
     In recent years, scholars have proposed various multi-label feature selection algorithms to address the curse of dimensionality and improve prediction accuracy~\cite{cai2011bassum,yu2020causality}. Multi-label feature selection algorithms are generally classified into four major categories~\cite{pereira2018categorizing}:  methods based on mutual information, mutual information-based, regularization-based, manifold learning-based, and causal structure learning-based methods. Methods based on mutual information, such as FIMF~\cite{lee2015fast}, SCLS~\cite{lee2017scls}, and SRLG-LMA~\cite{dai2024multi}, select the most relevant features by measuring the mutual information relationship between features and labels. Methods based on regularization, such as SFUS~\cite{ma2012web}, JFSC~\cite{huang2017joint}, and MLFS-GLOCAL~\cite{faraji2024multi}, constrain the complexity of the model by introducing regularization terms, thereby selecting the most important features. Methods based on manifold learning, such as MCLS~\cite{huang2018manifold}, MSSL~\cite{cai2018multi}, and MDFS~\cite{zhang2019manifold}, select features with important information by considering the local geometric structure and manifold characteristics of the data. Methods based on causal structure learning, such as MB-MCF~\cite{wu2020multi}, explore causal relationships between variables to construct a potential causal structure and select causally relevant features.
    
    Fan et al. proposed a new method called LCIFS~\cite{fan2024learning}, which integrates manifold learning, adaptive spectral graph, and redundancy analysis into an ensemble framework to learn relevant information for multi-label feature selection. LCIFS utilizes the structural correlation of labels and simultaneously controls the use of redundant features, thereby achieving multi-label feature selection with a clear objective function.
    
    
    In the near past, some scholars have focused on feature selection in federated learning. Federated learning can implement algorithms on multiple datasets while preserving privacy. Specifically, Hu et al. proposed a multi-participant federated evolutionary feature selection algorithm~\cite{hu2022multi} for imbalanced data under privacy protection. They introduced a multi-level joint sample filling strategy to address imbalanced or empty classes on each participant. Subsequently, a federated evolutionary feature selection method based on supervised particle swarm optimization with multiple participants was proposed by periodically sharing the optimal feature subset among participants. Banerjee et al. introduced an information-theoretic multi-label feature selection method called Fed-FiS~\cite{banerjee2021fed}. Fed-FiS evaluates feature-feature mutual information and feature-class mutual information to obtain local feature subsets and a global feature set.
    
    In summary, some feature selection algorithms have been proposed in federated environments, but they have not considered feature selection in multi-label scenarios and have not addressed causal relationships between labels and labels, labels and features, and features and features. Therefore, this paper proposes a novel causal multi-label feature selection algorithm considering data privacy issues.

\section{Notations and Definitions}\label{sec3}
    In this section, we initially present some key concepts and symbols related to causal structures and Bayesian networks in federated setting. We define $F=\{F_1,F_2,\ldots,F_m\}$ as a set of $m$ features, $Y=\{Y_1,Y_2,\ldots,Y_q\}$ as a set of $q$ labels, and $V=Y\cup F=\{V_1,V_2,\ldots,V_{m+q}\}$ as the node set encompassing all labels and features. Assuming that there are $N$ clients, with each client’s local data having a sample size of $W_n, n=1,2,\ldots,N$. During the learning process within the federated environment, the server sends a triplet $<Y_i,V_k,CS>$ to each client for performing the conditional independence test between $Y_i$ and $V_k$ under the conditioning set of $CS$ ($CS$ can be empty set $\emptyset$). Each client returns the correlation value $C_{n<V_i,V_k,CS>}$ and the P value $P_{n<V_i,V_k,CS>}$, enabling the server to aggregate the results into the corresponding weighted correlation value $C_{<V_i,V_k,CS>}$ and weighted P value $P_{<V_i,V_k,CS>}$. The significance level of the conditional independence test is denoted by $\alpha$ ($\alpha=0.05$). The notation $V_i\indep V_j\mid CS$ (where $i\neq j$ and $CS\subset V\setminus\{V_i,V_j\}$) denotes that $V_i$ is conditionally independent from $V_j$ given $CS$, while $V_i\nindep V_j\mid CS$ indicates that $V_i$ and $V_j$ are dependent given $CS$. The term $PC(Y_i)$ represents the set of local causal structure (parent-child node set) for the label $Y_i$.

    \begin{table}[htbp]
        \caption{Summary of Notations\label{tab1}}
        \centering
        \begin{tabular}{p{2cm}p{6cm}}
        \hline 
        Symbol   & Meaning \\
        \hline 
        $Y$ & a set of labels\\
        $F$ & a set of features\\
        $V$ & $V = Y \cup F$\\
        $q$ & the number of labels\\
        $m$ & the number of features\\
        $N$ & the number of clients\\
        $Y_i$ & a label in $Y$\\
        $F_i$ & a feature in $F$\\
        $V_i$ & a node in $V$\\
        $W_n$ & the sample size of the local data for the $n$th client ($n=1,2,\ldots,N$)\\
        $<Y_i,V_k,CS>$ & the triplet sent by the server\\
        $C_{n<V_i,V_k,CS>}$ & the correlation value between $Y_i$ and $V_k$ computed by the $n$th client under the $CS$ condition set\\
        $C_{<V_i,V_k,CS>}$ & the weighted correlation value\\
        $P_{n<V_i,V_k,CS>}$ & the P value between $Y_i$ and $V_k$ computed by the $n$th client under the $CS$ condition set\\
        $P_{<V_i,V_k,CS>}$ & the weighted P value\\
        $\alpha$ & the significance level of the CI test\\
        $V_i\indep V_j\mid CS$ & $V_i$ and $V_j$ are independent under the $CS$ condition set\\
        $V_i\nindep V_j\mid CS$ & $V_i$ and $V_j$ are dependent under the $CS$ condition set\\
        $PC (Y_i)$ & the parent and children set of the label variable $Y_i$\\
        \hline 
        \end{tabular}
    \end{table}

    \begin{definition}
    (\textbf{Bayesian Network, BN})~\cite{pearl1988probabilistic}. 
        Let $P(V)$ be the joint probability distribution over $V$ and $G=(V,E)$ represent a directed acyclic graph (DAG) with nodes $V$ and edges $E$. The triplet $<V,G,P(V)>$ is called a BN if and only if $<V,G,P(V)>$ satisfies the Markov condition: every node of $G$ is independent of any subset of its non-descendants conditioning on the parents of the node.
    \end{definition}

    In a DAG of BN, if there is a directed edge from A to B, A is the direct cause of B and  B is the direct effect of A, then the DAG is called a causal DAG (i.e. causal structure)~~\cite{pearl2009causality}.

    \begin{definition}
    (\textbf{Parent and Child, PC}~\cite{pearl2009causality}). 
        The parents and children of variable $V_i$ in a causal structure consists of the parents and children  of $V_i$, called $PC(V_i)$.
    \end{definition}
    
    \begin{definition}
    (\textbf{Markov blanket, MB}~\cite{pearl2009causality}). 
        The MB of a variable in a causal structure consists of the variable’s parents (direct causes), children (direct effects), and spouses (other parents of the variable's children).
    \end{definition}

    \begin{theorem}~\cite{pearl2009causality}\label{theorem3.1}
        In a DAG, given the MB of variable $V_i$,$MB(V_i)$, for $\forall V_j\in V\setminus(MB(V_i)\cup V_i)$, $V_i$ is conditionally independent of $V_j$ given $MB(V_i)$.	
    \end{theorem}

    Theorem~\ref{theorem3.1} indicates that in classification tasks, the MB of a label variable is the optimal feature subset for predicting the label variable~\cite{yu2021unified}. Furthermore, recent studies suggest that in real-world scenarios, the prediction quality of a label variable’s PC set is almost identical to that of the MB of the class variable~\cite{yu2020causality,aliferis2010local}. Therefore, we employ a well-established local causal structure learning algorithm, HITON-PC~\cite{aliferis2010local}, to learn the PC set of a class variable (any other state-of-the-art  local causal structure learning algorithms can be used here).

    \begin{figure}[htbp]
        \centering
        \includegraphics[width=2.3in]{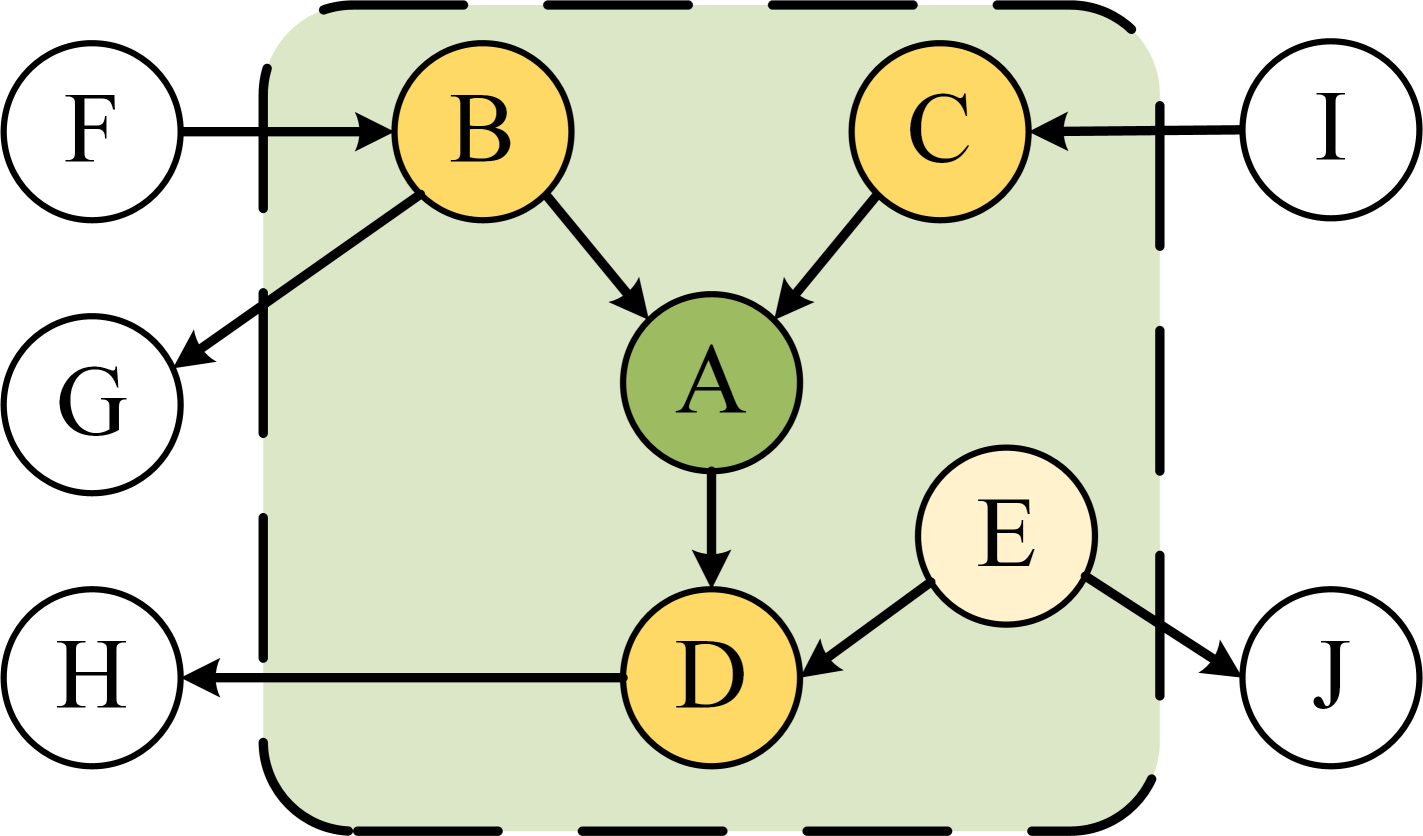}
        \caption{The MB of node A consisting of B,C,D,E. The PC set of node A comprising B, C, and D.}
        \label{BN_structure}
    \end{figure}
   
\section{Proposed FedCMFS Algorithm}\label{sec4}
\subsection{Overview of the FedCMFS Algorithm}\label{sec4_1}
    In this paper, we simulate a federated environment using a client-server architecture, and propose the FedCMFS algorithm for causal multi-label feature selection in federated setting. FedCMFS is a horizontal federated learning algorithm, which uses a central server and multiple clients to perform causal feature selection on standard multi-label data. Specifically, FedCMFS sequentially executes the following three subroutines to select causal features:
    (1) Federated Causal Feature Learning algorithm (FedCFL); (2) Federated Causal Feature Retrieval algorithm (FedCFR); and (3) Federated Causal Feature Correction algorithm (FedCFC).
    
    FedCFL treats both labels and features as ordinary variables, and it independently computes local causal variables for each label on each client. Throughout local causal structure learning, the clients  interact with the server, ultimately obtaining causally relevant feature sets for all label variables: $PC\left(Y\right) = \{CPC\left(Y_1\right),CPC\left(Y_2\right),...,CPC\left(Y_q\right)\}$ for all labels on the server.
    
    To tackle the issue of missing true relevant features, in FedCFR, the server communicate with each client to selectively identify potentially missing true relevant features, obtaining an updated causally relevant feature sets for all label variables: $PC_{SR} = \{PC\left (Y_1\right ), PC\left (Y_2\right ), \ldots, PC\left (Y_q\right )\}$.
    
    Due to the data quality issue, utilizing the symmetry property of causal neighbors, FedCFC selectively corrects false causally relevant features in  $PC_{SR}$ and achieves the final feature set $sel=PC\left (Y_1\right )\cup PC\left( Y_2\right )\cup\ldots\cup PC\left (Y_q\right )$.

\subsection{Federated Causal Feature Learning Algorithm (FedCFL)}\label{sec4_2}
    HITON-PC~\cite{aliferis2010local} is a widely used algorithm for learning a PC set of a given variable from a single-label dataset, which adopts a forward-backward strategy, exhibiting notable performance in causal feature learning. In this paper, we extend HITION-PC to the learn PC set of a label variable for multi-label data in federated setting and propose the federated causal feature learning algorithm, FedCFL, to address the causal multi-label feature selection problem in federated setting.
    
    A simple strategy for applying the HITON-PC algorithm to multi-label feature selection in federated setting is that each client learns a PC set independently for each label variable, and then aggregates the PC sets at the service. However, due to the different quality of samples from different clients, the PC sets of a given label variable learned from different clients are often different. To deal with this issue, we design the FedCFL algorithm consisting of two learning phases, as shown in Alg.~\ref{FedCFL}.

     \begin{algorithm}[H]
    \caption{FedCFL Algorithm.}\label{FedCFL}
    \begin{algorithmic}[1]
        \REQUIRE $Y$
        \ENSURE $PC(Y)$
        
        \emph{//Phase I: Identify potential PC variables of each label}
        \FOR{each client $n$}
        	\FORALL{$Y_i \in Y$}
        		\FORALL{$V_k \in V \setminus Y_i$}
        			\STATE Client $n$ calculate $C_{n<Y_i,V_k,\emptyset>}$ and $P_{n<Y_i,V_k,\emptyset>}$ and add them to $M_{n,i}^\prime$;
        		\ENDFOR
        	\ENDFOR
        	\STATE Client $n$ send $M_n^\prime$ to the server;
        \ENDFOR
        \STATE Server merges, prunes, and sorts $M^\prime$ to obtain $M$;
        
        \emph{//Phase II: Execute forward-backward strategy to update CPC($Y_i$)}
        \FORALL{$M_i \in M$}
            \FORALL{$V_j \in M_i$}
                \STATE Add $V_j$ to $CPC(Y_i)$; 
                \FORALL{$V_k \in CPC(Y_i)$}
                	\FORALL{$CS: CS \subseteq CPC(Y_i) \setminus V_k$}
                		\STATE Server sends $<Y_i,V_k,CS>$ to all clients;
                		\FOR{each client $n$}
                		\STATE Client $n$ calculate  $P_{n<Y_i,V_k,CS>}$ and send it to server;
                		\ENDFOR
                		\STATE Server computes $P_{<Y_i,V_k,CS>}$;
                		\IF{$P_{<Y_i,V_k,CS>} > \alpha$}
                			\STATE Server remove $V_k$ from $CPC(Y_i)$;
                			\STATE Break;
                		\ENDIF
                	\ENDFOR
                \ENDFOR
            \ENDFOR
        \ENDFOR
        
        \STATE $PC(Y) = \{CPC(Y_1), CPC(Y_2), \ldots, CPC(Y_q)\}$;
        \RETURN $PC(Y)$
    \end{algorithmic}
    \end{algorithm}

    In Phase I of FedCFL, we initially identify potential PC variables for each class label, where the variables contain both features and labels, so that correlations between not only label-features, feature-features, but also label-labels can be learned in the process of learning the local causal structure.  The computation for this phase is carried out at each client and ultimately converges at the server. Assuming that there are $N$ clients (denoted as client 1, client 2, …, client N) and one server. At the beginning of phase I, upon the server’s request, each client independently computes the correlation value $C_{n<Y_i,V_k,\emptyset>}$ and the P value $P_{n<Y_i,V_k,\emptyset>}$ between each class label $Y_i$ and the other variables $V_k\in V\setminus Y_i$ on a local client, using the empty set as the condition set via conditional independence tests. The node $V_k$, correlation value $C_{n<Y_i,V_k,\emptyset>}$ and P value $P_{n<Y_i,V_k,\emptyset>}$ are then added to the initial correlation set $M_{ni}^\prime$ (where $n$ represents the client number $n\ \in1,2,\ldots,N$, and $i$ represents the label number $i\ \in1,2,\ldots,q$). After computing the  correlations of all labels and all features, each client obtains a local initial correlation set $M_n^\prime=\{M_{n1}^\prime,M_{n2}^\prime,\ldots,M_{nq}^\prime\}$, which includes the initial correlation sets $M_{ni}^\prime$ for all class labels. Each client sends the learned local initial correlation set $M_n^\prime$ to the server. 

    \begin{figure}[htbp]
        \centering
        \includegraphics[width=\linewidth]{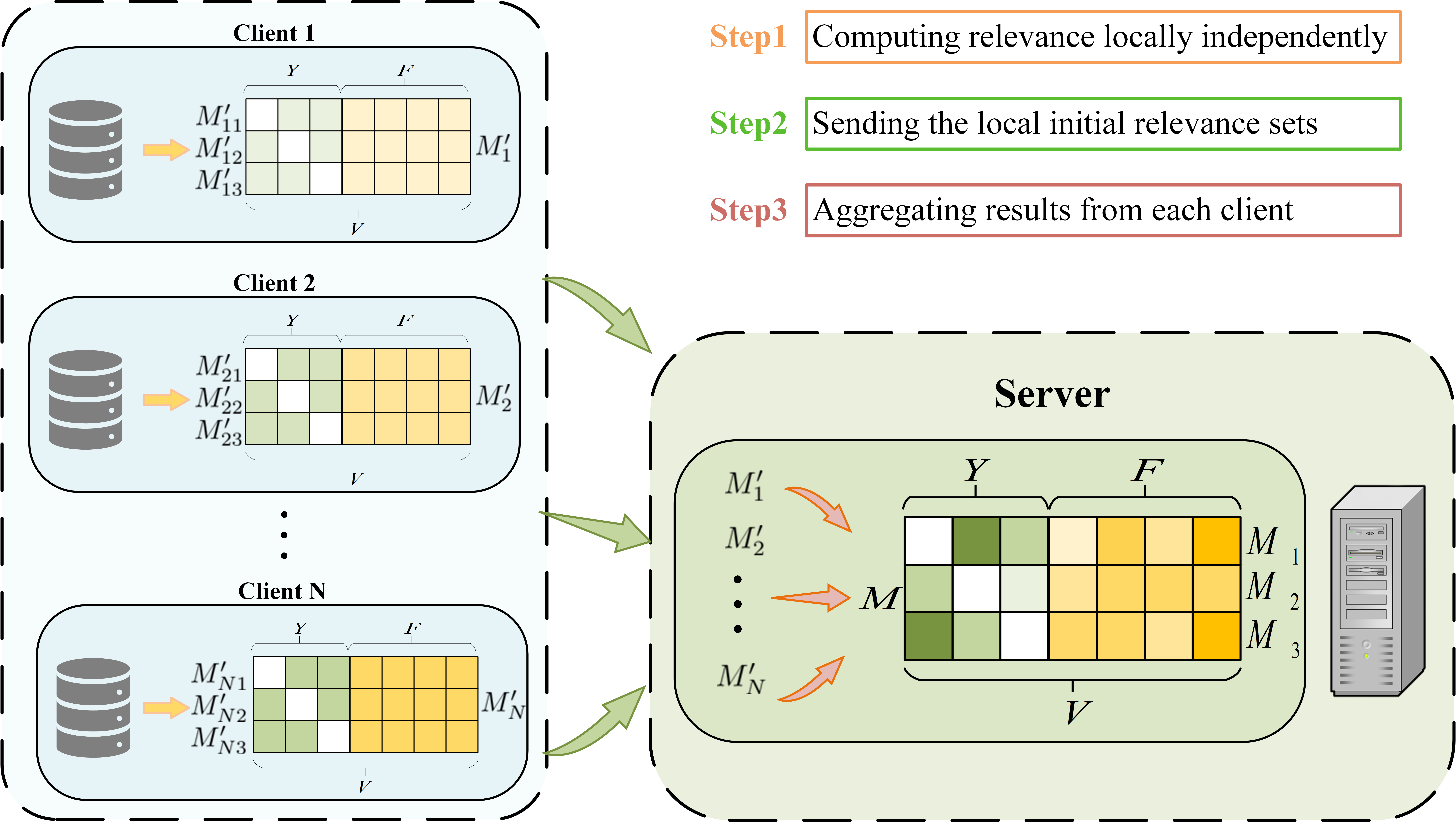}
        \caption{Phase I of FedCFL: Green and yellow are labels and features, respectively, and the figure shows the computation, transmission, and aggregation of a dataset containing three labels and four features. The absence of color and the presence of other colors in the squares signify that each client independently calculates the correlation value and P value between each label ($Y_i$) and all nodes excluding it ($V_k\in V\setminus Y_i$) on a local level.}
        \label{HITONPC_Phase1}
    \end{figure}

    The server receives the local initial correlation sets $M_1^\prime,M_2^\prime,\ldots,M_N^\prime$ learned by all clients, merges and prunes them according to Eq.~\ref{equation4-1} to obtain the global initial correlation set $M=\{M_1,M_2, \ldots,M_q\}$. Specifically, for each node $V_k$ that may be dependent with the target label $Y_i$, the server computes the weighted average $P_{<Y_i,V_k,\emptyset>}$ of the P value $P_{n<Y_i,V_k,\emptyset>}$ of the pair of variables $Y_i$ and $V_k$ across all clients. Assuming that a client with a large number of data samples may have a higher probability of representing the true statistical patterns, we assign the weight of each client, $W_n$, based on the proportion of data with regard to the total data samples across all clients. 

    \begin{equation}
        \begin{aligned}
        P_{<Y_{i},V_{k},CS>}=\frac{\sum_{n=1}^{N}\left(P_{n<Y_{i},V_{k},CS>}\cdot W_{n}\right)}{\sum_{n=1}^{N}W_{n}}
        \label{equation4-1}
        \end{aligned}
    \end{equation}

    If the weighted P value $P_{<Y_i,V_k,\emptyset>}$  is less than the significance level $\alpha$, the server determines that $Y_i$ and $V_k$ are not independent given the empty set as the condition set and computes the weighted average $C_{<Y_i,V_k,\emptyset>}$ of the correlation value $C_{n<Y_{i},V_{k},\emptyset>}$ of the pair of variables $Y_i$ and $V_k$ across all clients according to Eq.~\ref{equation4-2}. Subsequently, the server adds $C_{<Y_i,V_k,\emptyset>}$, $P_{<Y_i,V_k,\emptyset>}$, and $V_k$ to $M_i$.

    \begin{equation}
        \begin{aligned}
        C_{<Y_{i},V_{k},CS>}=\frac{\sum_{n=1}^{N}\left(C_{n<Y_{i},V_{k},CS>}\cdot W_{n}\right)}{\sum_{n=1}^{N}W_{n}}
        \label{equation4-2}
        \end{aligned}
    \end{equation} 

    After the global initial correlation set $M_i$ for each label is completed, the server sorts the variables in $M_i$ by the weighted correlation value $C_{<Y_i,V_k,\emptyset>}$ in descending order. Finally, the global initial correlation set $M=\{M_1,M_2,\ldots,M_q\}$ is obtained at the server. 

    In Phase II of FedCFL, we utilize a forward-backward strategy to progressively update the variables in the candidate parent and children set $CPC\left(Y_i\right)$ (initially $CPC\left(Y_i\right)$ is an empty set $\emptyset$) until a complete local causal structure is learned. To reduce the number of conditional independence tests, the computation performed by each client in Phase II is uniformly controlled by the server. The server sequentially adds the variable with the highest weighted correlation value in $M_i$, along with its corresponding weighted correlation value $C_{<Y_i,V_k,\emptyset>}$ and weighted P value $P_{<Y_i,V_k,\emptyset>}$, to the candidate set $CPC\left(Y_i\right)$. Whenever a new variable $X$ is added to $CPC\left(Y_i\right)$, the server needs to determine whether each variable $V_k$ in the current $CPC\left(Y_i\right)$ will be independent of the target label $Y_i$ under the condition that the new variable is added, and prune the $CPC\left(Y_i\right)$ based on the above result. Therefore, the server sends the triplet $<Y_i,V_k,CS>$ (where $Y_i$, $V_k$ are the variables to be tested, and $CS$ is the conditional set, $CS\subset CPC\left(Y_i\right)\setminus V_k)$ to all clients, requesting each to conduct the corresponding conditional independence tests.
    
    Once receiving the triplet $<Y_i,V_k,CS>$, each client independently conducts conditional independence tests between $Y_i$ and $V_k$ given the condition set $CS$, and subsequently returns the P value $P_{n<Y_i,V_k,CS>}$ between $Y_i$ and $V_k$ to the server. 
    
    Subsequently, the server receives the results from all clients and computes the weighted average $P_{<Y_i,V_k,CS>}$ of the P value $P_{n<Y_i,V_k,CS>}$ under $<Y_i,V_k,CS>$ across all clients according to Eq.~\ref{equation4-1} (the weight of each client is the proportion of data contained in that client out of the total data volume of all clients.). If the weighted P value $P_{<Y_i,V_k,CS>}$ exceeds the significance level $\alpha$, $Y_i$ is  independent of $V_k$, and $V_k$ along with its corresponding $C_{<Y_i,V_k,\emptyset>}$ and $P_{<Y_i,V_k,\emptyset>}$ are removed from $CPC\left(Y_i\right)$, otherwise they are retained.

    \begin{figure}[htbp]
        \centering
        \includegraphics[width=\linewidth]{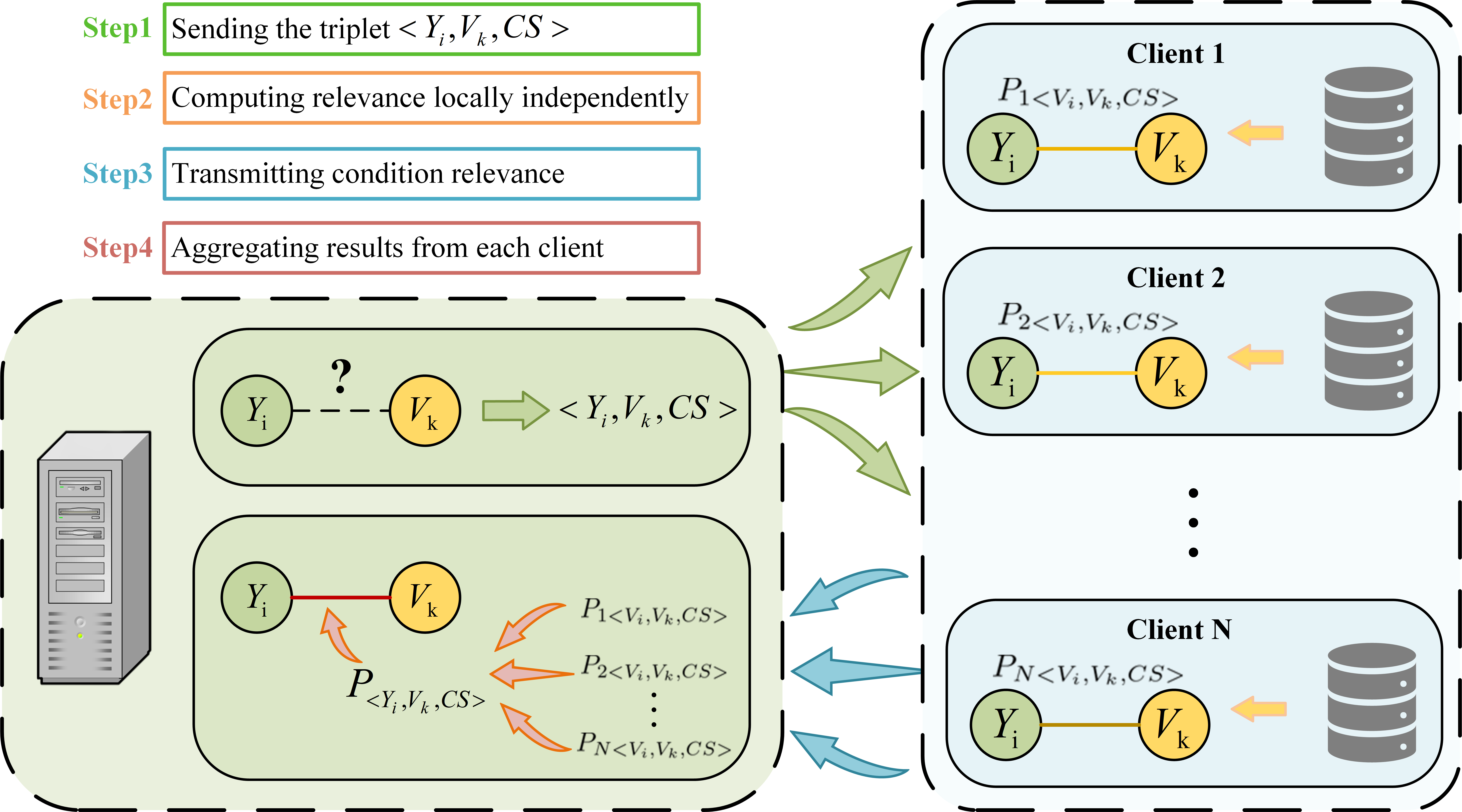}
        \caption{Phase II of FedCFL: the server sends triplet $<Y_i,V_k,CS>$ to determine conditional independence, each client computes and returns the P value $P_{n<Y_i,V_k,CS>}$, culminating in the aggregation of the weighted P value $P_{<Y_i,V_k,CS>}$.}
        \label{HITONPC_Phase2}
    \end{figure}

    The server and each client  interact to execute the aforementioned steps, and the server achieves the $PC$ set $PC\left(Y\right) = \{CPC\left(Y_1\right),CPC\left(Y_2\right),...,CPC\left(Y_q\right)\}$, which is a collection that not only contains the causal feature sets of each class label, but also includes the weighted correlation of the variables with their corresponding labels with an empty conditioning set. The completion of this step signifies the ending of the FedCFL algorithm.

\subsection{Federated Causal Feature Retrieval Algorithm (FedCFR)}\label{sec4_3}
    In FedCFL, all labels and features are treated as ordinary variables to simultaneously consider three types of correlations among variables in a multi-label dataset: feature-label, feature-feature, and label-label correlations. However, due to the correlation among labels, some true PC features may become independent of the labels, resulting in missing true PC features.

    \begin{figure}[htbp]
       \centering
        \includegraphics[width=2.3in]{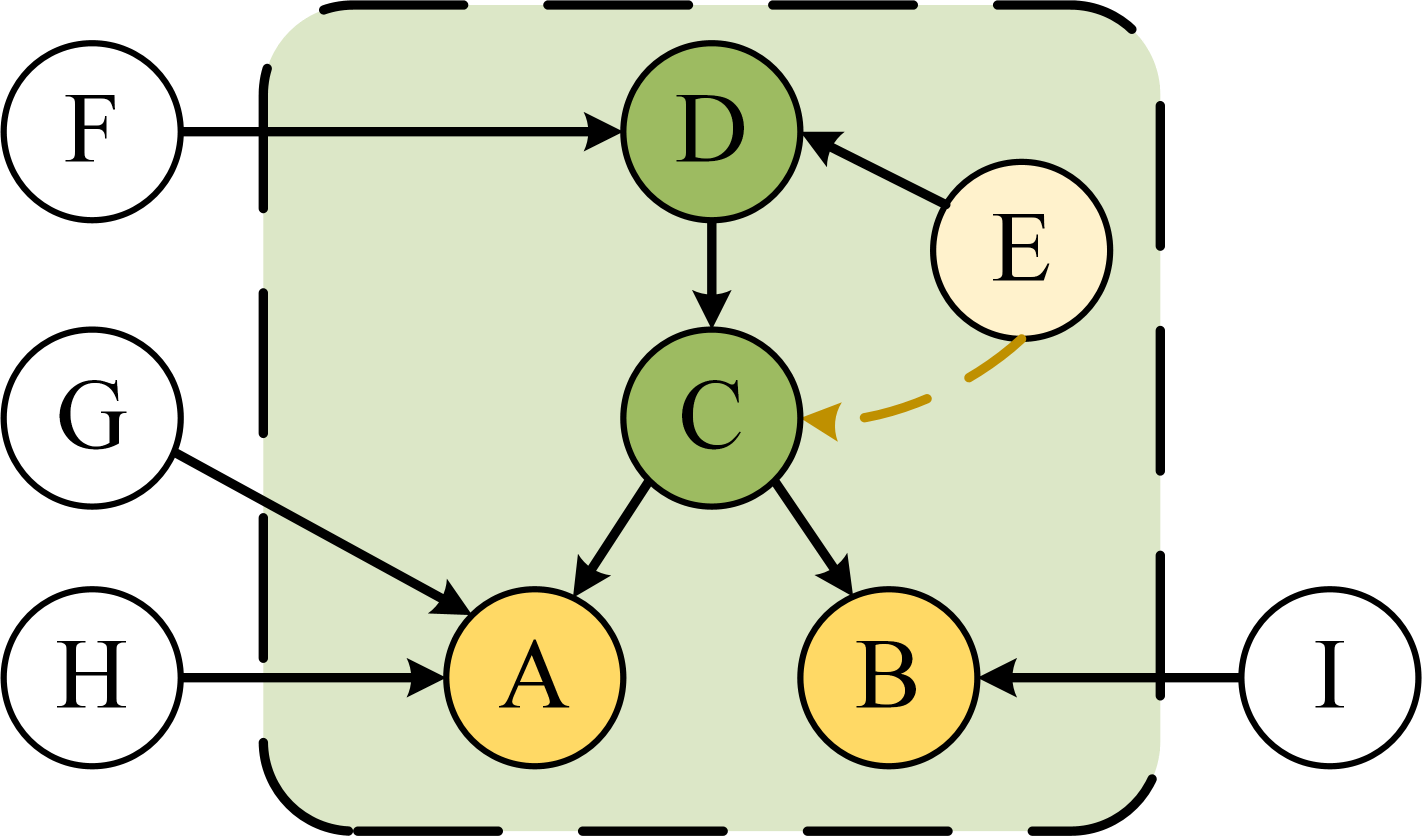}
       \caption{The label-label correlations lead to missing true causal features.}
        \label{Missing_feature}
    \end{figure}

    Taking the structure shown in the Figure~\ref{Missing_feature} as an example, suppose the PC set of the class label C consists of ${A,B,E}$. Due to the correlation between class label C and class label D, feature E, which serves as a parent node of the class label C, is not selected to the PC set of C. Therefore, to tackle this issue, we design the Federated Causal Feature Retrieval algorithm (FedCFR) and its pseudocodes are shown in Alg.~\ref{FedCFR}.

    The initial step of Phase I in FedCFR involves identifying potential missed PC features from all discarded features. From Steps 3 to 15, the FedCFR subroutine determines whether each label $Y_i\in Y$ needs to selectively retrieve missed PC features by judging whether its $PC\left(Y_i\right)$ contains other class labels. If $PC\left(Y_i\right)$ contains the label $Y_k$ ($k=1,\ldots,q$ and $k \neq i$), the possible missed PC features are searched for in discarded variables other than the selected features (i.e., $F_j\in F\setminus PC\left(Y_i\right)$). The judgment rule is: if $F_j\in F\setminus PC\left(Y_i\right)$ and $F_j\nindep Y_i\mid\emptyset$, it is considered to be a possible missed PC feature and added to the candidate feature set $sel\left(Y_i\right)$. 

        \begin{algorithm}[htbp]
    \caption{FedCFR Algorithm.}\label{FedCFR}
    \begin{algorithmic}[1]
        \REQUIRE $q$, $k_1$
        \ENSURE $PC_{SR}$
        
        \emph{//Phase I: Identify potential true PC features}
        \FOR{$i = 1$ \textbf{to} $q$}
            \IF{there exists $k$ in $1$ to $q$ such that $k \neq i$ and $Y_k \in PC(Y_i)$}
                \FORALL{$F_j \in F \setminus PC(Y_i)$}
                	\STATE Server sends $<Y_i,F_j,\emptyset>$ to all clients;
                	\FOR{each client $n$}
                	\STATE Client $n$ calculate  $P_{n<Y_i,F_j,\emptyset>}$, $C_{n<Y_i, F_j, \emptyset>}$ and send them to server;
                	\ENDFOR
                	\STATE Server computes $P_{<Y_i,F_j,\emptyset>}$, $C_{<Y_i, F_j, \emptyset>}$;
                	\IF{$P_{<Y_i,F_j,\emptyset>}  >\alpha$}
                	\STATE Add $F_j$, $C_{<Y_i, F_j, \emptyset>}$ and $P_{<Y_i, F_j, \emptyset>}$ to $sel(Y_i)$;
                	\ENDIF
                \ENDFOR
                \STATE $sel(Y_i) \leftarrow$ Sort features in $sel(Y_i)$ in descending order, take the top $k_1\%$ features;
            \ENDIF
        \ENDFOR
        
        \emph{//Phase II: Select the correct PC features from the potential sel$(Y_i)$}
        \FOR{$i = 1$ \textbf{to} $q$}
            \FOR{$j = 1$ \textbf{to} $q$}
                \IF{$Y_j \in PC(Y_i)$}
                    \FORALL{$X \in sel(Y_i)$}
                        \FORALL{$S: \{Y_j\} \subset S \subset PC(Y_i)$}
                        	\FORALL{$CS \subseteq CPC(Y_i) \setminus V_k$}
                        		\STATE Server sends $<X,Y_i,S>$,$<X,Y_i,S\setminus Y_j>$ to all clients;
                        		\FOR{each client $n$}
                        			\STATE Client $n$ calculate  $P_{n}$,  $C_{n}$, and send them to server;
                        		\ENDFOR
                        		\STATE Server computes $P_{n}$,  $C_{n}$;
                        		\IF{$P_{n<X,Y_i,S>} >\alpha$ and $P_{n<X,Y_i,S\setminus Y_j>} <\alpha$}
                        			\STATE Add $X$, $C_{<X, Y_i, \emptyset>}$ and $P_{<X, Y_i, \emptyset>}$ to $PC(Y_i)$;
                        		\ENDIF
                        	\ENDFOR
                        \ENDFOR
                    \ENDFOR
                    \STATE Remove $Y_j$, $C_{<Y_i, Y_j, \emptyset>}$ and $P_{<Y_i, Y_j, \emptyset>}$ from $PC(Y_i)$ ;  
                \ENDIF  
            \ENDFOR
        \ENDFOR
        
        \STATE $PC_{SR} = \{PC(Y_1), PC(Y_2), \ldots, PC(Y_q)\}$;
        \RETURN $PC_{SR}$
    \end{algorithmic}
    \end{algorithm}

    To aggregate the conditional independence test results among all clients in federated setting, the server sequentially sends the triplet $<Y_i,F_j,\emptyset>$ to all clients, requesting each client to perform the conditional independence tests with the empty set as the conditioning set and returning their results $C_{n<Y_i,F_j,\emptyset>}$ and $P_{n<Y_i,F_j,\emptyset>}$. Upon receiving the returned results from all clients, the server performs the calculations according to Eq.~\ref{equation4-1} and Eq.~\ref{equation4-2}. If the weighted average of the returned results from all clients $P_{<Y_i,F_j,\emptyset>}$ is less than the significance level $\alpha$, it is considered dependent. The variable and its corresponding $C_{<Y_i,F_j,\emptyset>}$ and $P_{<Y_i,F_j,\emptyset>}$, are then added to $sel\left(Y_i\right)$. Otherwise, it is considered independent.

    Recent research has shown that a causal structure in real-world scenarios is often relatively sparse~\cite{friedman2000using,peng2009partial}. For instance, in a dataset containing 1000 features, a class variable may only have 10 PC features. This implies that the discarded set $sel\left(Y\right)$ contains very few missed PC features. If all variables in $sel\left(Y\right)$ are tested, it would cost a significant amount of time. Among those candidate variables, the variables with high correlations to a class variable are most likely to be the missed PC features. Thus in Step 8, FedCFR addresses this issue by sorting the candidate feature set in descending order according to the weighted correlation value $C_{<Y_i,F_j,\emptyset>}$ with the empty conditioning set, then selects the top $k_1$\% variables.

    In the second phase, FedCFR utilizes the available structural information to determine candidate features that may have been missed by FedCFL. Taking Figure~\ref{Missing_feature} as an example, supposing the learned PC set of label C is $PC(C)=\{A,B,D\}$, and E has not been correctly added to the PC set. After the first phase of FedCFR, feature E is added to the candidate feature set $sel(C)$. In this case, there must exist a set $S$ containing the class label D such that $C\indep E\mid S$. If the class label D is removed from $S$, C will be dependent of E.
    
   Therefore, the server traverses $\forall Y_i,Y_j\in Y$, and when class label $Y_j$ appears in $PC\left(Y_i\right)$ of the class label $Y_i$, it examines the top $k_1$\% variables in the candidate feature set $sel\left(Y_i\right)$. If a set $S: Y_j\subset S\subset PC\left(Y_i\right)$ is found such that $X\in sel\left(Y_i\right)$ and $X  \indep Y_i\mid S$, and $X \nindep Y_i\mid S\setminus Y_j$, then $X$ is considered a missed PC feature. $X$ and its corresponding $C_{<X,Y_i,\emptyset>}$ and $P_{<X,Y_i,\emptyset>}$ are added to $PC\left(Y_i\right)$. 

    To coordinate all clients to complete the above operations in federated setting, the server sends the triplet $<X,Y_i,S>$ to all clients and receives the results returned by each client. It then uses Eq.~\ref{equation4-1} to obtain the weighted P value $P_{<X,F_j,S>}$. If the weighted P value  $P_{<X,F_j,S>}$ exceeds the significance level $\alpha$, it is considered independent. The server then sends the triplet $<X,Y_i,S\setminus Y_j>$ to each client and confirms whether the $P_{<X,F_j,S\setminus Y_j>}$ results are dependent. After the successful completion of both tests, $X$, $C_{<X,Y_i,\emptyset>}$ and $P_{<X,Y_i,\emptyset>}$ are added to $PC\left(Y_i\right)$. When all operations against $Y_j$ in $sel\left(Y_i\right)$ are completed, the label $Y_j$ is removed from $PC\left(Y_i\right)$. After the execution of the FedCFR algorithm, the set of the missed PC features is obtained, denoted as $PC_{SR} =\{PC\left(Y_1\right),PC\left(Y_2\right),\ldots,PC\left(Y_q\right)\}$. 

\subsection{Federated Causal Feature Correction Algorithm (FedCFC)}\label{sec4_4}
    In the first two subroutines of FedCMFS, the uneven quality of data across clients may lead to erroneous aggregation results using the conditional independence tests. It means that both FedCFL and FedCFR may learn false PC variables. To remove these false PC variables in  $PC\left(Y_i\right)$ (obtained by FedCFL and FedCFR), by the  symmetry property of a parent and its children in a DAG (if  A is a parent of  B, then B must be a child of A), we design the FedCFC subroutine which  examines whether the PC of a variable in $PC\left(Y_i\right)$ includes $Y_i$, to eliminate false PC features in $PC\left(Y_i\right)$. The pseudocodes of the FedCFC algorithm is provided below.

    \begin{algorithm}[H]
        \caption{FedCFC Algorithm.}\label{FedCFC}
        \begin{algorithmic}[1]
        \REQUIRE $q$, $k_2$
        \ENSURE $sel$
        
        \FOR{$i = 1$ \textbf{to} $q$}
            \STATE $CanF(Y_i) \leftarrow $ Sort the features in $PC(Y_i)$ in ascending order of the relevance, take the top $k_2\%$ features;
            \FORALL{$F_j \in CanF(Y_i)$}
                \STATE $PC\left(F_j\right) = FedCFL(F_j)$;
                \IF{$Y_i \notin PC(F_j)$}
                    \STATE Remove $F_j$, $C_{<Y_i, F_j, \emptyset>}$ and $P_{<Y_i, F_j, \emptyset>}$ from $PC(Y_i)$;  
                \ENDIF
            \ENDFOR
        \ENDFOR
        
        \STATE $sel = PC(Y_1)\cup PC(Y_2)\cup \ldots \cup PC(Y_q)$;
        \RETURN $sel$
        \end{algorithmic}
    \end{algorithm}

    Given that the true $PC\left(Y_i\right)$ contains fewer false features, the features with the smaller correlation to the class variable are more likely to be mistakenly selected. To avoid calculating the PC of all variables in $PC\left(Y_i\right)$, the server initially sorts the weighted correlation value with an empty set and store them in ascending order. It then selects the top $k_2\%$ features with the smallest correlations to the class label $Y_i$ to be included in the correction set $CanF\left(Y_i\right)$. 
    
    In Steps 3 to 9, the FedCFC Algorithm applies the FedCFL algorithm to the feature in  $F_j\in CanF\left(Y_i\right)$ and obtains the result at the  server. It then determines whether the set $PC\left(F_j\right)$ of the feature $F_j$ contains the label $Y_i$. If not,  $F_j$ is removed from the set $PC\left(Y_i\right)$ of $Y_i$. After the correction process for all labels is completed, the finally selected feature set $sel=PC\left(Y_1\right)\cup PC\left(Y_2\right)\cup\ldots\cup PC\left(Y_q\right)$ is obtained.

\subsection{Acceleration Method}\label{sec4_5}
    The three subroutines of FedCMFS all involve conditional independence tests (CI tests). Thus the time complexity of FedCMFS can be measured by the number of CI tests performed by a single client. The time complexity of FedCMFS is relatively high, with the time complexity of the subroutine FedCFL being $O((m+q)q2^{|PC(V)|})$, the time complexity of the subroutine FedCFR being $ O(mq^22^{|PC(V)|}) $, and the time complexity of the subroutine FedCFC being $ O(|PC|mq2^{|PC(V)|}) $. This indicates that the number of CI tests performed is exponentially related to the size of the learned PC set. As the dimensionality of the dataset increases, the learned PC set may become larger, and the number of tests correspondingly increases. Therefore, to improve the execution speed of FedCMFS, this paper proposes the following three strategies to accelerate CI tests by leveraging the parallel computing ability of GPU.
    
    The first strategy involves data-level parallel processing for two types of CI tests: the G² test~\cite{neapolitan2004learning} for discrete data and the Fisher's Z test~\cite{neapolitan2004learning} for continuous data. A brief introduction to these two CI testing methods is provided below.
    
    Assuming that variables $X_i$ and $X_j$ are conditionally independent given $X_k$, the formula for the G² test algorithm is as follows:

    \begin{equation}
    	\begin{aligned}
    		\mathrm{G}^2=2\sum_{a,b,c}s_{ijk}^{abc}\ln\left(\frac{s_{ijk}^{abc}s_k^c}{s_{ik}^{ac}s_{jk}^{bc}}\right)
    		\label{equation4-3}
    	\end{aligned}
    \end{equation}
    
    $G^2$ is a statistic, $S_{ijk}^{abc}$ is a random variable, whose value is the number of times $X_i=a$, $X_j=b$, and $X_k=c$ simultaneously occur in a data sample. The null hypothesis of independence is rejected by calculating the p-value of the $G^2$ statistic and comparing it to a predetermined level of significance.
    
    Assuming that variables $X_i$ and $X_j$ are conditionally independent given $S$, the formula for the Fisher's Z test algorithm is as follows:
    
    \begin{equation}
    	\begin{aligned}
    		Z=\frac{1}{2} \sqrt{M-\mid S\mid-3}\biggl(\ln\frac{1+R}{1-R}\biggr)
    		\label{equation4-4}
    	\end{aligned}
    \end{equation}
    
    $Z$ is a statistic, $M$ is the sample size, and $R$ denotes a random variable whose value is the partial correlation coefficient of $X_i$ and $X_j$ given the condition set $S$. The null hypothesis of independence is rejected by calculating the p-value of the $Z$ statistic and comparing it to a predetermined level of significance.
    
	GPUs have many processing cores that can handle multiple data streams simultaneously, enabling massively parallel computing. In this paper, we fully utilize the parallel computing capabilities of GPUs to implement the parallel computation of the G² test and Fisher's Z test. Next, we will take the calculation of $S_{ijk}^{abc}$  as an example to explain in detail.

    When calculating $S_{ijk}^{abc}$, for each sample in the dataset, all possible combinations of $S_{ijk}$ are traversed to determine whether \(X_i = a\), \(X_j = b\), and \(X_k = c\) are satisfied, and the count is updated accordingly. Assuming \(X_i\), \(X_j\), and \(X_k\) each have 100 values, in the worst case, \(10^6\) conditional judgments need to be performed to ensure that all possible combinations are considered.
	
	On the GPU, \(S_{ijk}\) is organized as a three-dimensional tensor, with each dimension corresponding to \(X_i\), \(X_j\), and \(X_k\) respectively. For the dimension of the variable \(X_i\), the task of finding an index with the value \(a\) is assigned in parallel to multiple processing units. Each processing unit is responsible for determining whether a value in the \(X_i\) dimension is equal to \(a\), and if so, returning the index. Similarly, the indexes with values \(b\) and \(c\) are found in the \(X_j\) and \(X_k\) dimensions using the same method. After obtaining the indexes, \(S_{ijk}^{abc}\) is updated by incrementing the corresponding count. Using this parallelization method, only three parallel operations on the GPU are required to perform a task that would otherwise take \(10^6\) operations on the CPU, resulting in a significant increase in computational speed.
	
	The second strategy is the vectorized parallelization of Fisher's Z test. Function vectorization involves applying a function to each element of a vector, allowing multiple data elements to be processed simultaneously. For example, consider the simple function \( f(x) = x^2 \). If this function is vectorized, then for a vector \( x \) containing multiple elements, the square of each element can be computed directly without the need for an explicit loop to compute them one by one.
	
	In this paper, Fisher's Z test is treated as a function and its vectorized parallelism is implemented on GPUs. When dealing with continuous data, the execution process of the FedCMFS algorithm is optimized such that the server sends all CI test requests of the current phase to the client in batches. The client then executes all test requests in parallel, instead of using an explicit loop to perform the tests sequentially, and returns all the results to the server at once. Next, the first phase of the FedCFL subroutine is described in detail as an example.
	
	In the first phase of FedCFL, the server sends requests to the clients to perform tests on \(Y_i\) and \(V_k\) under the empty set (\(\emptyset\)) one at a time. Clients perform these tests sequentially, with each client needing to conduct \(q \times (m + q - 1)\) CI tests.
	
	After using the vectorized parallelization of Fisher's Z test, as shown in Algorithm \ref{accFedCFL}, the server records all triples \((Y_i, V_k, \emptyset)\) into \(vecinput\) and sends \(vecinput\) to all clients. Upon receiving it, clients perform CI tests in batches. The tests within the same batch are executed in parallel on the GPU, and the results are recorded into \(vecoutput\). Subsequently, clients extract \(C_{n<Y_i,V_k,\emptyset>}\) and \(P_{n<Y_i,V_k,\emptyset>}\) from \(vecoutput\) and add them to \(M'_{ni}\). Finally, clients send \(M'_n\) back to the server. The batch size needs to be set according to the GPU's performance. If the batch size is set to 100, each client only needs to execute \(q \times (m+q-1)/100\) batches of CI tests.
 
    \begin{algorithm}
        \caption{Accelerated FedCFL Algorithm Phase I.}
        \label{accFedCFL}
        \begin{algorithmic}[1]
            \REQUIRE $Y$
            \ENSURE $M_n^\prime$
            \FOR{each client $n$}
            \STATE $vecinput=\emptyset$;
            \FORALL{$Y_i \in Y$}
            \FORALL{$V_k \in V \setminus Y_i$}
            \STATE Record $(Y, V_i, \emptyset)$ into $vecinput$;
            \ENDFOR
            \ENDFOR
            \STATE $vecoutput=vec\_fishers\_z\_test(vecinput)$;
            \FORALL{$Y_i \in Y$}
            \FORALL{$V_k \in V \setminus Y_i$}
            \STATE Take $C_{n<Y_i,V_k,\emptyset>}$, $P_{n<Y_i,V_k,\emptyset>}$ from $vecoutput$ and add them to $M_{n,i}^\prime$;
            \ENDFOR
            \ENDFOR  
            \STATE Send $M_n^\prime$ to the server;
            \ENDFOR
            
        \end{algorithmic}
    \end{algorithm}
	
	The third strategy is the CI test recording mechanism. Sometimes, two variables under the same condition set have already been tested, but a duplicate test is performed in a subsequent run, resulting in a waste of computational resources. We record the results of each test in the CI test record. Before executing a test, we first check the CI test record to see if a corresponding record already exists. If it does, the result in the record is used directly without repeating the test; if not, the test continues to be executed. This mechanism effectively reduces the number of CI tests and accelerates the algorithm.  
 
\section{Experiment}\label{sec5}
\subsection{Datasets}\label{sec5_1}
    To evaluate the effectiveness of FedCMFS, we use eight real-datasets to simulate the federated setting. The specific information of the eight  real-datasets is shown in Table~\ref{tab2}.

    \begin{table}[htbp]
        \caption{Description of datasets used in the experiments\label{tab2}}
        \setlength{\tabcolsep}{3.5pt}
        \centering
        \begin{tabular}{cccccc}
        \hline 
        Datasets &Instances  &Label &Feature &Domain &Data type\\
        \hline 
        CHD\_49     & 555   & 6       & 49     & Medicine   & Continuous \\
        VirusGo     & 207   & 6       & 749    & Biology    & Discrete \\
        Yeast       & 2417  & 14      & 103    & Biology    & Continuous \\
        Flags       & 194   & 7       & 19     & Image      & Discrete \\
        Image       & 2000  & 5       & 294    & Image      & Continuous \\
        Slashdot    & 3782  & 22      & 1079   & Text       & Discrete \\
        Business    & 5000  & 30      & 438    & Text       & Continuous \\
        Education   & 5000  & 33      & 550    & Text       & Continuous \\
        \hline 
        \end{tabular}
    \end{table}

\subsection{Evaluation Metrics}\label{sec5_2}
    We select six commonly used multi-label metrics to evaluate the performance of the methods, including Average precision (AP), Coverage (CV), Hamming loss (HL), Rank loss (RL), Macro-F1 (Fma), and Micro-F1 (Fmi). Suppose there is a multi-labeled dataset $D=\{\left(f_i,Y_i\mid1\le i\le s\right)\}$, where $f_i$ and $Y_i$ are the feature set and label set corresponding to the current $i$th sample, respectively. The ranking function $rank_i\left(y\right)$ denotes the predicted ranking of the $i$th sample corresponding to label $y$. The detailed explanation of the six evaluation metrics is as follows:

    (1) Average precision: the average of the scores of the correct labels for evaluating label alignment, where $AP\left(D\right)\in\left[0,1\right]$, the higher value the better performance is.
    \begin{equation}
        \begin{aligned}
        AP(D) &= \frac{1}{s}\sum_{i=1}^{s}\frac{1}{|Y_{i}|}\cdot\sum_{y\in Y_{i}} \\
        &\quad \frac{sum\bigl(rank_{i}(y^{*})\le rank_{i}(y)\bigr),y^{*}\in Y_{i}}{rank_{i}(y)}
        \label{equation5-1}
        \end{aligned}
    \end{equation}

    (2) Coverage: the average of the number of steps required by the sample to traverse all labels, the smaller value of $CV\left(D\right)$ the performance is.
    \begin{equation}
        \begin{aligned}
        CV(D)=\frac{1}{s}\sum_{i=1}^{s}maxrank_{i}(y)-1
        \label{equation5-2}
        \end{aligned}
    \end{equation}

    (3) Hamming loss: it is used to evaluate the proportion of samples that are incorrectly matched. Here, $q$ stands for the number of labels, $Z_i$ is the predicted label set, $Y_i$ represents the true label set of the current $i$th sample, and $\Delta$ is the symmetric difference between the predicted and the true label sets. $HL\left(D\right)\in\left[0,1\right]$, the smaller value the better performance is.
        \begin{equation}
            \begin{aligned}
            HL(D)=\frac{1}{s}\sum_{i=1}^{s}\frac{1}{\mathrm{q}}|Z_{i}\bigtriangleup Y_{i}|
            \label{equation5-3}
            \end{aligned}
        \end{equation}

    (4) Ranking Loss: evaluate the ranking of relevant labels over irrelevant labels for the samples, $RL\left(D\right)\in\left[0,1\right]$, the smaller value the better performance is.
    \begin{equation}
        \begin{aligned}
        RL(D)=\frac{1}{s}\sum_{i=1}^{s}\frac{1}{|Y_{i}||\overline{Y}_{i}|} \cdot|rank_{i}(y^{*})>rank(y^{**})|\\ y^{*}\in Y_{i},y^{**}\in\overline{Y}_{i}
        \label{equation5-4}
        \end{aligned}
    \end{equation}

    (5)Macro-F1: arithmetic mean of F1 scores.
    \begin{equation}
        \begin{aligned}
        Macro-F1=\frac{1}{s}\sum_{i=1}^{s}\frac{2TP_{i}}{2TP_{i}+FP_{i}+TN_{i}}
        \label{equation5-5}
        \end{aligned}
    \end{equation}

    (6)Micro-F1: weighted average of F1 scores.
        \begin{equation}
            \begin{aligned}
            Micro-F1=\frac{\sum_{i=1}^{s}2TP_{i}}{\sum_{i=1}^{s}2TP_{i}+FP_{i}+FN_{i}}
            \label{equation5-6}
            \end{aligned}
        \end{equation} 

    The two metrics mentioned above consider both the recall and precision of the model. Here, $TP_i$ represents the number of true positives in the model’s predictions, while $FP_i$ and $NP_i$ represent the number of false positives and false negatives, respectively. $Macro-F1, Micro-F1 \in\left[0,1\right]$, and a larger value indicates better performance.

\subsection{Comparison Algorithms}\label{sec5_3}
    There are no existing multi-label feature selection methods designed in federated environments for preserving data privacy. To validate our FedCMFS, we select five state-of-the-art feature selection methods, MB-MCF (Markov blanket-based multi-label causal feature selection)~\cite{wu2020multi}, GLFS (Group-preserving label-specific feature selection for multi-label learning)~\cite{zhang2023group}, PDMFS (Parallel Dual-channel Multi-label Feature Selection)~\cite{miao2023parallel}, GRROOR (Global Redundancy and Relevance Optimization in Orthogonal Regression for Embedded Multi-label Feature Selection)~\cite{zhang2020multi}, and PMFS (Pareto-based feature selection algorithm for multi-label classification)~\cite{hashemi2021efficient}, and then we adopt a simple strategy to make the above five comparison algorithms work in federated setting. Specifically, for each comparison algorithm, each client independently executes the algorithm locally. Once all clients have completed their execution, they send the evaluation metric results to the server. The server then calculates a weighted average (with weights being the proportion of the data contained in each client to the total data volume) and records the weighted average of the evaluation metrics from all clients as the final result for comparison.

\subsection{Experimental Environment and Parameter Settings}\label{sec5_4}
    All experiments were conducted on a Ubuntu server equipped with an Intel(R) Xeon(R) Platinum 8375C CPU @2.90GHz CPU, 64GB of memory, and an NVIDIA A100-SXM 40GB GPU. The programming environment utilized was Python 3.8, with PyTorch version 2.1.0. The batch size was set to 100.

    For FedCMFS algorithm proposed in this paper, parameters $k_1$ and $k_2$ for FedCFR and FedCFC are set within $\left(0,0.3\right]$. The significance level for conditional independence tests is $\alpha=0.05$. All other comparison algorithms used default parameter settings. 

    The ML-KNN~\cite{zhang2007ml} method is chosen for the experiments to evaluate the results of federated feature selection, which is a widely adopted classifier in multi-label learning(any other multi-label classification algorithms can be used here), and the parameter $k$ of the algorithm  is set to 10 according to the default setting. Among all the algorithms, only MB-MCF algorithm selects a fixed number of features, while other comparison algorithms select features randomly. To effectively track metric variations in the final feature subset, all algorithms, except for MB-MCF, align with the FedCMFS algorithm using the identical number of features for federated causal multi-label feature selection.

    To simulate the federated setting, this study assumes a total of $N$ clients, where $N\ \in\{3,5,10\ \}$, each with $W_n$ data items. The data division method for the simulated federated setting is as follows: (1) For smaller datasets: CHD\_49, VirusGo, Yeast, Flags, and Image, each client randomly extracts $40\%-60\%$ of the original training set data from the real dataset without repetition to construct their dataset. (2) For larger datasets: Slashdot, Business, and Education, each client randomly extracts $30\%-50\%$ of the original training set data from the real dataset without repetition to construct their dataset. Although the data within a client is unique, there may be data overlap among different clients.

\subsection{Experimental Results and Analysis}\label{sec5_5}
    The comparison results of the six algorithms, FedCMFS, MB-MCF, GLFS, PDMFS, GRROOR, and PMFS are shown in Tables~\ref{tab3} to \ref{tab8}. In the table, the optimal results are shown in bold, with higher ↑ values being better, lower ↓ values being better, and \emph{Average} representing the average ranking of the current method among the six methods.
    
     \begin{table}[htbp]
        \caption{Experimental result of Average Precision (↑).\label{tab3}}
        \setlength{\tabcolsep}{3.0pt}
        \scriptsize 
        \centering
        \begin{tabular}{m{0.84cm}<{\centering}m{0.83cm}<{\centering}m{0.89cm}<{\centering}m{1cm}<{\centering}m{0.89cm}<{\centering}m{0.89cm}<{\centering}m{0.89cm}<{\centering}m{0.89cm}<{\centering}}
        \hline 
        Datasets    &Clients  &FedCMFS    &MB-MCF &GLFS   &PDMFS  &GRROOR &PMFS\\
        \hline 
        \multirow{3}{*}{CHD\_49}  & 3  & \textbf{0.7722} & 0.7708 & 0.7707 & 0.7537 & 0.7634 & 0.7693 \\
                                  & 5  & \textbf{0.7783} & 0.7647 & 0.7671 & 0.7577 & 0.7567 & 0.7608 \\
                                  & 10 & \textbf{0.7732} & 0.7664 & 0.7685 & 0.7525 & 0.7571 & 0.7607 \\ 
        \hline
        \multirow{3}{*}{VirusGo}  & 3  & \textbf{0.9452} & 0.9432 & 0.6712 & 0.6333 & 0.6340 & 0.6279 \\
                                  & 5  & \textbf{0.9442} & 0.9341 & 0.6516 & 0.6354 & 0.6330 & 0.6321 \\
                                  & 10 & \textbf{0.9442} & 0.9353 & 0.6620 & 0.6334 & 0.6320 & 0.6291 \\ 
        \hline
        \multirow{3}{*}{Yeast}    & 3  & 0.7571 & 0.7503 & \textbf{0.8210} & 0.7306 & 0.7395 & 0.8156 \\
                                  & 5  & 0.7542 & 0.7523 & 0.8170 & 0.7304 & 0.7340 & \textbf{0.8210} \\
                                  & 10 & 0.7590 & 0.7535 & 0.8172 & 0.7286 & 0.7364 & \textbf{0.8190} \\ 
        \hline
        \multirow{3}{*}{Flags}    & 3  & 0.8251 & 0.8113 & 0.8283 & \textbf{0.8534} & 0.7709 & 0.7719 \\
                                  & 5  & 0.8224 & 0.8019 & 0.8195 & \textbf{0.8476} & 0.7899 & 0.7786 \\
                                  & 10 & 0.7856 & 0.7992 & \textbf{0.8397} & 0.8343 & 0.7850 & 0.7812 \\ 
        \hline
        \multirow{3}{*}{Image}    & 3  & \textbf{0.7199} & 0.6893 & 0.5688 & 0.5277 & 0.6771 & 0.6585 \\
                                  & 5  & \textbf{0.7133} & 0.6838 & 0.5657 & 0.5386 & 0.6993 & 0.7025 \\
                                  & 10 & \textbf{0.7247} & 0.6883 & 0.5303 & 0.5296 & 0.6415 & 0.6911 \\ 
        \hline
        \multirow{3}{*}{Slashdot} & 3  & \textbf{0.7477} & 0.7465 & 0.7413 & 0.7397 & 0.7411 & 0.7398 \\
                                  & 5  & \textbf{0.7480} & 0.7452 & 0.7406 & 0.7397 & 0.7399 & 0.7427 \\
                                  & 10 & \textbf{0.7461} & 0.7457 & 0.7399 & 0.7388 & 0.7401 & 0.7414 \\ 
        \hline
        \multirow{3}{*}{Business} & 3  & \textbf{0.8793} & 0.8759 & 0.8692 & 0.8787 & 0.8688 & 0.8743 \\
                                  & 5  & \textbf{0.8798} & 0.8768 & 0.8690 & 0.8733 & 0.8688 & 0.8719 \\
                                  & 10 & 0.8769 & 0.8770 & 0.8719 & \textbf{0.8806} & 0.8682 & 0.8700 \\ 
        \hline
        \multirow{3}{*}{Education}& 3  & \textbf{0.5735} & 0.5668 & 0.5436 & 0.5622 & 0.5239 & 0.5345 \\
                                  & 5  & \textbf{0.5705} & 0.5704 & 0.5376 & 0.5354 & 0.5278 & 0.5295 \\
                                  & 10 & \textbf{0.5763} & 0.5696 & 0.5427 & 0.5317 & 0.5316 & 0.5323 \\ 
        \hline
        \emph{Average}   & $\backslash$ & \textbf{1.5833} & 2.7083 & 3.2500 & 4.4167 & 5.0000 & 4.0417 \\ 
        \hline 
        \end{tabular}
    \end{table}

    \begin{table}[htbp]
        \caption{Experimental result of Coverage (↓).\label{tab4}}
        \setlength{\tabcolsep}{3.0pt}
        \scriptsize 
        \centering
        \begin{tabular}{m{0.84cm}<{\centering}m{0.83cm}<{\centering}m{0.89cm}<{\centering}m{1cm}<{\centering}m{0.89cm}<{\centering}m{0.89cm}<{\centering}m{0.89cm}<{\centering}m{0.89cm}<{\centering}}
        \hline 
        Datasets    &Clients  &FedCMFS    &MB-MCF &GLFS   &PDMFS  &GRROOR &PMFS\\
        \hline 
       \multirow{3}{*}{CHD\_49}   & 3  & \textbf{0.4702} & 0.4891 & 0.4911 & 0.5069 & 0.5010 & 0.5089 \\
                                  & 5  & \textbf{0.4702} & 0.4940 & 0.4952 & 0.5238 & 0.5119 & 0.5101 \\
                                  & 10 & \textbf{0.488}1 & 0.4917 & 0.4908 & 0.5205 & 0.5199 & 0.5137 \\ 
        \hline
        \multirow{3}{*}{VirusGo}  & 3  & \textbf{0.0703} & 0.0730 & 0.2296 & 0.2450 & 0.2456 & 0.2423 \\
                                  & 5  & \textbf{0.0622} & 0.0719 & 0.2373 & 0.2450 & 0.2442 & 0.2390 \\
                                  & 10 & \textbf{0.0622} & 0.0735 & 0.2301 & 0.2454 & 0.2452 & 0.2428 \\ 
        \hline
        \multirow{3}{*}{Yeast}    & 3  & 0.4642 & 0.4631 & \textbf{0.4081} & 0.4775 & 0.4722 & 0.4128 \\
                                  & 5  & 0.4618 & 0.4623 & \textbf{0.4122} & 0.4783 & 0.4761 & 0.4127 \\
                                  & 10 & 0.4531 & 0.4604 & \textbf{0.4122} & 0.4804 & 0.4730 & 0.4129 \\ 
        \hline
        \multirow{3}{*}{Flags}    & 3  & \textbf{0.5165} & 0.5253 & 0.5465 & 0.5289 & 0.5568 & 0.5802 \\
                                  & 5  & \textbf{0.5077} & 0.5543 & 0.5464 & 0.5323 & 0.5508 & 0.5824 \\
                                  & 10 & 0.5407 & 0.5512 & \textbf{0.5251} & 0.5391 & 0.5637 & 0.5635 \\ 
        \hline
        \multirow{3}{*}{Image}    & 3  & \textbf{0.2180} & 0.2257 & 0.3523 & 0.4133 & 0.2490 & 0.2767 \\
                                  & 5  & \textbf{0.2185} & 0.2293 & 0.3602 & 0.3993 & 0.2285 & 0.2358 \\
                                  & 10 & \textbf{0.2130} & 0.2298 & 0.3905 & 0.4065 & 0.2635 & 0.2439 \\ 
        \hline
        \multirow{3}{*}{Slashdot} & 3  & \textbf{0.0346} & 0.0347 & 0.0396 & 0.0401 & 0.0388 & 0.0400 \\
                                  & 5  & \textbf{0.0349} & 0.0354 & 0.0400 & 0.0410 & 0.0395 & 0.0390 \\
                                  & 10 & 0.0365 & \textbf{0.0348} & 0.0405 & 0.0411 & 0.0403 & 0.0394 \\ 
        \hline
        \multirow{3}{*}{Business} & 3  & \textbf{0.0761} & 0.0767 & 0.0798 & 0.0768 & 0.0800 & 0.0788 \\
                                  & 5  & \textbf{0.0756} & 0.0770 & 0.0804 & 0.0780 & 0.0805 & 0.0803 \\
                                  & 10 & \textbf{0.0765} & 0.0772 & 0.0795 & 0.0766 & 0.0804 & 0.0798 \\ 
        \hline
        \multirow{3}{*}{Education}& 3  & \textbf{0.1130} & 0.1138 & 0.1206 & 0.1158 & 0.1248 & 0.1207 \\
                                  & 5  & 0.1145 & \textbf{0.1139} & 0.1199 & 0.1221 & 0.1230 & 0.1218 \\
                                  & 10 & \textbf{0.1132} & 0.1139 & 0.1199 & 0.1241 & 0.1224 & 0.1216 \\ 
        \hline
        \emph{Average}   & $\backslash$ & \textbf{1.4583} & 2.4583 & 3.3750 & 4.8750 & 4.7917 & 4.0417 \\ 
        \hline 
        \end{tabular}
    \end{table}

    \begin{table}[htbp]
        \caption{Experimental result of Hamming loss (↓).\label{tab5}}
        \setlength{\tabcolsep}{3.0pt}
        \scriptsize 
        \centering
        \begin{tabular}{m{0.84cm}<{\centering}m{0.83cm}<{\centering}m{0.89cm}<{\centering}m{1cm}<{\centering}m{0.89cm}<{\centering}m{0.89cm}<{\centering}m{0.89cm}<{\centering}m{0.89cm}<{\centering}}
        \hline 
        Datasets    &Clients  &FedCMFS    &MB-MCF &GLFS   &PDMFS  &GRROOR &PMFS\\
        \hline 
       \multirow{3}{*}{CHD\_49}  & 3  & \textbf{0.3006} & 0.3214 & 0.3284 & 0.3333 & 0.3284 & 0.3284 \\
                                  & 5  & \textbf{0.3095} & 0.3173 & \textbf{0.3095} & 0.3494 & 0.3405 & 0.3363 \\
                                  & 10 & \textbf{0.3065} & 0.3235 & 0.3193 & 0.3443 & 0.3339 & 0.3366 \\ 
        \hline
        \multirow{3}{*}{VirusGo}  & 3  & \textbf{0.0462} & 0.0469 & 0.2001 & 0.1988 & 0.1988 & 0.2001 \\
                                  & 5  & \textbf{0.0402} & 0.0498 & 0.1996 & 0.1988 & 0.1988 & 0.2000 \\
                                  & 10 & \textbf{0.0402} & 0.0528 & 0.1990 & 0.1988 & 0.1988 & 0.1990 \\ 
        \hline
        \multirow{3}{*}{Yeast}    & 3  & 0.1985 & 0.2010 & \textbf{0.1692} & 0.2155 & 0.2099 & 0.1724 \\
                                  & 5  & 0.2038 & 0.2019 & \textbf{0.1721} & 0.2148 & 0.2125 & 0.1726 \\
                                  & 10 & 0.1982 & 0.2012 & 0.1724 & 0.2157 & 0.2116 & \textbf{0.1721} \\ 
        \hline
        \multirow{3}{*}{Flags}    & 3  & \textbf{0.2857} & 0.2872 & 0.2989 & 0.3011 & 0.3267 & 0.3480 \\
                                  & 5  & 0.2989 & 0.3116 & 0.3059 & 0.3059 & \textbf{0.2941} & 0.3323 \\
                                  & 10 & 0.3165 & 0.3046 & \textbf{0.2958} & 0.3066 & 0.3198 & 0.3345 \\ 
        \hline
        \multirow{3}{*}{Image}    & 3  & 0.2065 & \textbf{0.2015} & 0.2320 & 0.2312 & 0.2117 & \textbf{0.2015} \\
                                  & 5  & 0.1995 & 0.2191 & 0.2255 & 0.2288 & 0.2009 & \textbf{0.1892} \\
                                  & 10 & 0.2145 & 0.2147 & 0.2321 & 0.2327 & 0.2296 & \textbf{0.1935} \\ 
        \hline
        \multirow{3}{*}{Slashdot} & 3  & \textbf{0.0177} & 0.0179 & 0.0249 & 0.0215 & 0.0202 & 0.0262 \\
                                  & 5  & 0.0183 & \textbf{0.0179} & 0.0217 & 0.0215 & 0.0209 & 0.0270 \\
                                  & 10 & 0.0177 & \textbf{0.0177} & 0.0215 & 0.0213 & 0.0215 & 0.0232 \\ 
        \hline
        \multirow{3}{*}{Business} & 3  & 0.0267 & 0.0273 & 0.0283 & \textbf{0.0265} & 0.0281 & 0.0266 \\
                                  & 5  & \textbf{0.0270} & 0.0272 & 0.0283 & 0.0274 & 0.0283 & 0.0275 \\
                                  & 10 & 0.0271 & 0.0272 & 0.0277 & \textbf{0.0264} & 0.0284 & 0.0278 \\ 
        \hline
        \multirow{3}{*}{Education}& 3  & 0.0400 & 0.0402 & 0.0418 & \textbf{0.0399} & 0.0426 & 0.0422 \\
                                  & 5  & \textbf{0.0400} & 0.0404 & 0.0422 & 0.0420 & 0.0422 & 0.0426 \\
                                  & 10 & \textbf{0.0401} & 0.0404 & 0.0419 & 0.0417 & 0.0425 & 0.0426 \\ 
        \hline
        \emph{Average}   & $\backslash$ & \textbf{1.8750} & 2.6250 & 3.8958 & 4.0417 & 4.2917 & 4.2708 \\
        \hline 
        \end{tabular}
    \end{table}

    \begin{table}[htbp]
        \caption{Experimental result of Ranking loss (↓).\label{tab6}}
        \setlength{\tabcolsep}{3.0pt}
        \scriptsize 
        \centering
        \begin{tabular}{m{0.84cm}<{\centering}m{0.83cm}<{\centering}m{0.89cm}<{\centering}m{1cm}<{\centering}m{0.89cm}<{\centering}m{0.89cm}<{\centering}m{0.89cm}<{\centering}m{0.89cm}<{\centering}}
        \hline 
        Datasets    &Clients  &FedCMFS    &MB-MCF &GLFS   &PDMFS  &GRROOR &PMFS\\
        \hline 
       \multirow{3}{*}{CHD\_49}  & 3  & \textbf{0.2277} & 0.2369 & 0.2373 & 0.2547 & 0.2473 & 0.2459 \\
                                  & 5  & \textbf{0.2177} & 0.2442 & 0.2393 & 0.2603 & 0.2544 & 0.2532 \\
                                  & 10 & \textbf{0.2369} & 0.2452 & 0.2384 & 0.2623 & 0.2588 & 0.2505 \\ 
        \hline
        \multirow{3}{*}{VirusGo}  & 3  & \textbf{0.0429} & 0.0449 & 0.2324 & 0.2536 & 0.2544 & 0.2506 \\
                                  & 5  & \textbf{0.0352} & 0.0445 & 0.2426 & 0.2536 & 0.2526 & 0.2464 \\
                                  & 10 & \textbf{0.0352} & 0.0457 & 0.2336 & 0.2541 & 0.2538 & 0.2509 \\ 
        \hline
        \multirow{3}{*}{Yeast}    & 3  & 0.1761 & 0.1781 & \textbf{0.1284} & 0.1929 & 0.1869 & 0.1332 \\
                                  & 5  & 0.1773 & 0.1763 & 0.1318 & 0.1946 & 0.1901 & \textbf{0.1300} \\
                                  & 10 & 0.1717 & 0.1757 & \textbf{0.1318} & 0.1966 & 0.1889 & 0.1318 \\ 
        \hline
        \multirow{3}{*}{Flags}    & 3  & 0.2087 & 0.2126 & 0.2079 & \textbf{0.1828} & 0.2535 & 0.2648 \\
                                  & 5  & 0.1910 & 0.2325 & 0.2159 & \textbf{0.1894} & 0.2402 & 0.2571 \\
                                  & 10 & 0.2354 & 0.2368 & \textbf{0.1945} & 0.2027 & 0.2479 & 0.2459 \\ 
        \hline
        \multirow{3}{*}{Image}    & 3  & \textbf{0.2304} & 0.2394 & 0.3958 & 0.4702 & 0.2685 & 0.3021 \\
                                  & 5  & \textbf{0.2281} & 0.2445 & 0.4039 & 0.4520 & 0.2429 & 0.2518 \\
                                  & 10 & \textbf{0.2254} & 0.2451 & 0.4432 & 0.4612 & 0.2883 & 0.2608 \\ 
        \hline
        \multirow{3}{*}{Slashdot} & 3  & 0.0381 & \textbf{0.0374} & 0.0427 & 0.0430 & 0.0423 & 0.0431 \\
                                  & 5  & \textbf{0.0375} & 0.0382 & 0.0427 & 0.0432 & 0.0424 & 0.0422 \\
                                  & 10 & 0.0395 & \textbf{0.0377} & 0.0428 & 0.0435 & 0.0428 & 0.0423 \\ 
        \hline
        \multirow{3}{*}{Business} & 3  & \textbf{0.0403} & 0.0413 & 0.0432 & 0.0404 & 0.0436 & 0.0422 \\
                                  & 5  & \textbf{0.0397} & 0.0411 & 0.0439 & 0.0418 & 0.0438 & 0.0434 \\
                                  & 10 & 0.0406 & 0.0412 & 0.0431 & \textbf{0.0403} & 0.0441 & 0.0433 \\ 
        \hline
        \multirow{3}{*}{Education}& 3  & \textbf{0.0867} & 0.0873 & 0.0936 & 0.0890 & 0.0978 & 0.0936 \\
                                  & 5  & 0.0876 & \textbf{0.0873} & 0.0931 & 0.0953 & 0.0964 & 0.0951 \\
                                  & 10 & \textbf{0.0866} & 0.0874 & 0.0932 & 0.0968 & 0.0955 & 0.0949 \\ 
        \hline
        \emph{Average}   & $\backslash$ & \textbf{1.6667} & 2.5417 & 3.3542 & 4.6667 & 4.8542 & 3.9167 \\ 
        \hline 
        \end{tabular}
    \end{table}

    \begin{table}[htbp]
        \caption{Experimental result of Macro-F1 (↑).\label{tab7}}
        \setlength{\tabcolsep}{3.0pt}
        \scriptsize 
        \centering
        \begin{tabular}{m{0.84cm}<{\centering}m{0.83cm}<{\centering}m{0.89cm}<{\centering}m{1cm}<{\centering}m{0.89cm}<{\centering}m{0.89cm}<{\centering}m{0.89cm}<{\centering}m{0.89cm}<{\centering}}
        \hline 
        Datasets    &Clients  &FedCMFS    &MB-MCF &GLFS   &PDMFS  &GRROOR &PMFS\\
        \hline 
       \multirow{3}{*}{CHD\_49}  & 3  & \textbf{0.4736} & 0.4245 & 0.3956 & 0.3473 & 0.4007 & 0.3564 \\
                                  & 5  & \textbf{0.4855} & 0.4418 & 0.4287 & 0.3908 & 0.3403 & 0.3851 \\
                                  & 10 & \textbf{0.4619} & 0.4277 & 0.4310 & 0.4199 & 0.3543 & 0.3711 \\ 
        \hline
        \multirow{3}{*}{VirusGo}  & 3  & 0.5862 & \textbf{0.6125} & 0.0000 & 0.0108 & 0.0217 & 0.0000 \\
                                  & 5  & \textbf{0.6743} & 0.6124 & 0.0000 & 0.0195 & 0.0000 & 0.0000 \\
                                  & 10 & \textbf{0.6743} & 0.5477 & 0.0103 & 0.0130 & 0.0065 & 0.0000 \\ 
        \hline
        \multirow{3}{*}{Yeast}    & 3  & 0.3545 & 0.3472 & \textbf{0.4237} & 0.2755 & 0.3080 & 0.4017 \\
                                  & 5  & 0.3508 & 0.3467 & 0.3961 & 0.2759 & 0.2911 & \textbf{0.4052} \\
                                  & 10 & 0.3530 & 0.3472 & 0.3957 & 0.2699 & 0.2949 & \textbf{0.4066} \\ 
        \hline
        \multirow{3}{*}{Flags}    & 3  & \textbf{0.5470} & 0.4900 & 0.4449 & 0.4785 & 0.4944 & 0.4431 \\
                                  & 5  & \textbf{0.5619} & 0.4700 & 0.4349 & 0.4608 & 0.4937 & 0.4290 \\
                                  & 10 & 0.3951 & \textbf{0.4792} & 0.4628 & 0.4502 & 0.4768 & 0.4312 \\ 
        \hline
        \multirow{3}{*}{Image}    & 3  & \textbf{0.4598} & 0.4418 & 0.1491 & 0.1448 & 0.4346 & 0.3731 \\
                                  & 5  & \textbf{0.5032} & 0.4334 & 0.1825 & 0.1362 & 0.4441 & 0.4375 \\
                                  & 10 & \textbf{0.4593} & 0.4300 & 0.1508 & 0.1501 & 0.3769 & 0.4352 \\ 
        \hline
        \multirow{3}{*}{Slashdot} & 3  & \textbf{0.0833} & 0.0795 & 0.0356 & 0.0393 & 0.0408 & 0.0357 \\
                                  & 5  & 0.0514 & \textbf{0.0856} & 0.0387 & 0.0392 & 0.0395 & 0.0333 \\
                                  & 10 & 0.0679 & \textbf{0.0873} & 0.0389 & 0.0393 & 0.0393 & 0.0374 \\ 
        \hline
        \multirow{3}{*}{Business} & 3  & 0.0941 & \textbf{0.0943} & 0.0641 & 0.0741 & 0.0644 & 0.0630 \\
                                  & 5  & \textbf{0.0998} & 0.0954 & 0.0623 & 0.0681 & 0.0647 & 0.0549 \\
                                  & 10 & \textbf{0.0991} & 0.0984 & 0.0658 & 0.0727 & 0.0583 & 0.0624 \\ 
        \hline
        \multirow{3}{*}{Education}& 3  & \textbf{0.0780} & 0.0744 & 0.0594 & 0.0571 & 0.0395 & 0.0511 \\
                                  & 5  & 0.0749 & \textbf{0.0773} & 0.0559 & 0.0387 & 0.0464 & 0.0503 \\
                                  & 10 & \textbf{0.0830} & 0.0759 & 0.0587 & 0.0399 & 0.0475 & 0.0495 \\ 
        \hline
        \emph{Average}   & $\backslash$ & \textbf{1.6667} & 2.2500 & 3.9792 & 4.5208 & 4.0625 & 4.5208 \\ 
        \hline 
        \end{tabular}
    \end{table}

    \begin{table}[htbp]
        \caption{Experimental result of Micro-F1 (↑).\label{tab8}}
        \setlength{\tabcolsep}{3.0pt}
        \scriptsize 
        \centering
        \begin{tabular}{m{0.84cm}<{\centering}m{0.83cm}<{\centering}m{0.89cm}<{\centering}m{1cm}<{\centering}m{0.89cm}<{\centering}m{0.89cm}<{\centering}m{0.89cm}<{\centering}m{0.89cm}<{\centering}}
        \hline 
        Datasets    &Clients  &FedCMFS    &MB-MCF &GLFS   &PDMFS  &GRROOR &PMFS\\
        \hline 
       \multirow{3}{*}{CHD\_49}  & 3  & \textbf{0.6599} & 0.6242 & 0.5993 & 0.5988 & 0.6217 & 0.5926 \\
                                  & 5  & \textbf{0.6623} & 0.6258 & 0.6368 & 0.6075 & 0.5837 & 0.6060 \\
                                  & 10 & \textbf{0.6532} & 0.6193 & 0.6250 & 0.6313 & 0.5965 & 0.6059 \\ 
        \hline
        \multirow{3}{*}{VirusGo}  & 3  & \textbf{0.8757} & 0.8746 & 0.0000 & 0.0249 & 0.0498 & 0.0000 \\
                                  & 5  & \textbf{0.8958} & 0.8684 & 0.0000 & 0.0449 & 0.0000 & 0.0000 \\
                                  & 10 & \textbf{0.8958} & 0.8580 & 0.0143 & 0.0299 & 0.0150 & 0.0000 \\ 
        \hline
        \multirow{3}{*}{Yeast}    & 3  & 0.6316 & 0.6275 & \textbf{0.6998} & 0.5804 & 0.5997 & 0.6891 \\
                                  & 5  & 0.6276 & 0.6239 & \textbf{0.6887} & 0.5819 & 0.5917 & 0.6877 \\
                                  & 10 & 0.6287 & 0.6274 & 0.6870 & 0.5766 & 0.5922 & \textbf{0.6898} \\ 
        \hline
        \multirow{3}{*}{Flags}    & 3  & \textbf{0.7032} & 0.6749 & 0.6583 & 0.6774 & 0.6444 & 0.6132 \\
                                  & 5  & \textbf{0.6909} & 0.6604 & 0.6452 & 0.6636 & 0.6756 & 0.6438 \\
                                  & 10 & 0.6230 & \textbf{0.6673} & 0.6637 & 0.6631 & 0.6534 & 0.6326 \\ 
        \hline
        \multirow{3}{*}{Image}    & 3  & \textbf{0.4643} & 0.4450 & 0.1522 & 0.1637 & 0.4378 & 0.4106 \\
                                  & 5  & \textbf{0.5006} & 0.4156 & 0.1925 & 0.1525 & 0.4688 & 0.4918 \\
                                  & 10 & 0.4211 & 0.4237 & 0.1583 & 0.1644 & 0.3718 & \textbf{0.4717} \\ 
        \hline
        \multirow{3}{*}{Slashdot} & 3  & \textbf{0.7848} & 0.7793 & 0.6750 & 0.7628 & 0.7722 & 0.6476 \\
                                  & 5  & 0.7807 &\textbf{ 0.7819} & 0.7464 & 0.7619 & 0.7638 & 0.6257 \\
                                  & 10 & \textbf{0.7864} & 0.7853 & 0.7502 & 0.7627 & 0.7617 & 0.7197 \\ 
        \hline
        \multirow{3}{*}{Business} & 3  & \textbf{0.7028} & 0.6954 & 0.6760 & 0.6943 & 0.6804 & 0.6910 \\
                                  & 5  & \textbf{0.7011} & 0.6957 & 0.6761 & 0.6847 & 0.6774 & 0.6812 \\
                                  & 10 & \textbf{0.7007} & 0.6970 & 0.6810 & 0.6944 & 0.6757 & 0.6807 \\ 
        \hline
        \multirow{3}{*}{Education}& 3  & \textbf{0.2501} & 0.2442 & 0.1594 & 0.2144 & 0.1168 & 0.1479 \\
                                  & 5  & \textbf{0.2454} & 0.2350 & 0.1479 & 0.1217 & 0.1320 & 0.1368 \\
                                  & 10 & \textbf{0.2582} & 0.2394 & 0.1657 & 0.1315 & 0.1291 & 0.1339 \\ 
        \hline
        \emph{Average}   & $\backslash$ & \textbf{1.5833} & 2.5000 & 3.9792 & 4.0417 & 4.4167 & 4.3958 \\ 
        \hline 
        \end{tabular}
    \end{table}

    The purpose of these experiments is to evaluate the performance of the FedCMFS algorithm on various metrics and datasets. The results show that FedCMFS consistently maintains the highest average ranking across all six evaluation metrics. In most cases, its performance is very close to that of the MB-MCF algorithm. Although FedCMFS performs slightly worse on the Yeast and Flags datasets, ranking in the top three, it performs well on high-dimensional datasets such as Slashdot, Business, and Education. This change in performance can be attributed to the effect of data sparsity on the Yeast and Flags datasets. This sparsity hinders the accuracy of statistical methods used for conditional independence tests, affecting the ability of FedCMFS to accurately learn causal structure. In contrast, in high-dimensional datasets, FedCMFS effectively identifies three different types of correlations between labels and features, demonstrating its superior performance.

\subsection{Parameter Sensitivity Analysis}\label{sec5_6}
\begin{figure*}[htb]
        \centering
        \subfloat[\footnotesize Parameter sensitivity analysis of FedCMFS on the Flags dataset when the number of clients is 3.]{\includegraphics[width=0.8\linewidth]{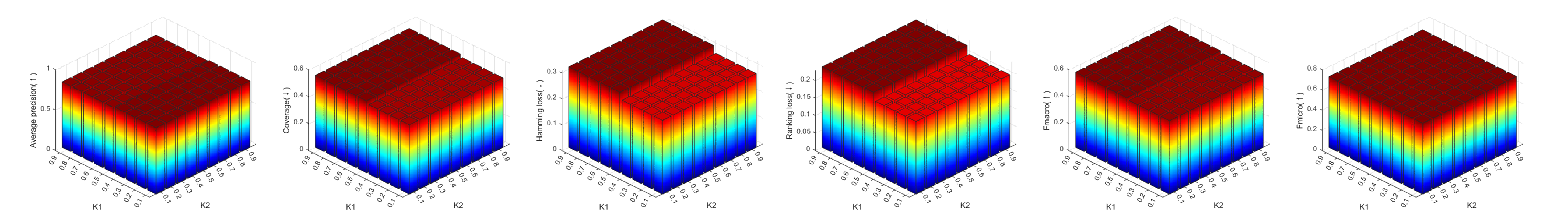}}
        \hfil
        \subfloat[\footnotesize Parameter sensitivity analysis of FedCMFS on the Flags dataset when the number of clients is 5.]{\includegraphics[width=0.8\linewidth]{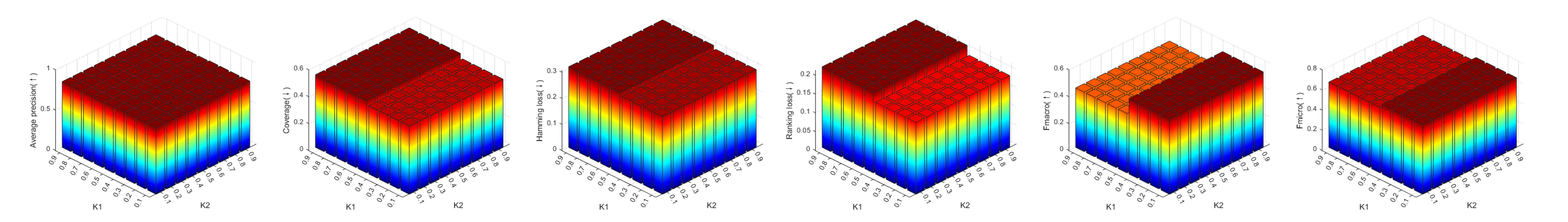}}
        \hfil
        \subfloat[\footnotesize Parameter sensitivity analysis of FedCMFS on the Flags dataset when the number of clients is 10.]{\includegraphics[width=0.8\linewidth]{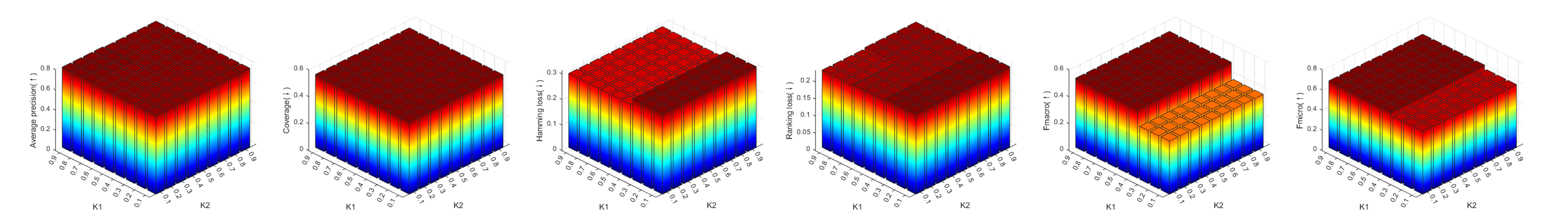}}
        \hfil
        \subfloat[\footnotesize Parameter sensitivity analysis of FedCMFS on the Image dataset when the number of clients is 3.]
        {\includegraphics[width=0.8\linewidth]{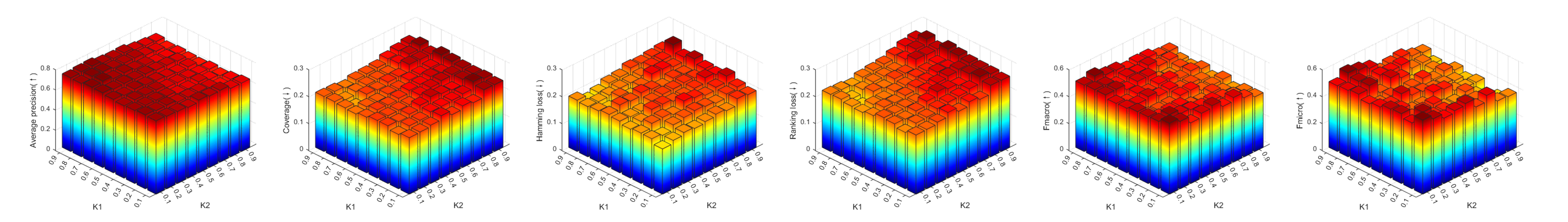}}
        \hfil
        \subfloat[\footnotesize Parameter sensitivity analysis of FedCMFS on the Image dataset when the number of clients is 5.]{\includegraphics[width=0.8\linewidth]{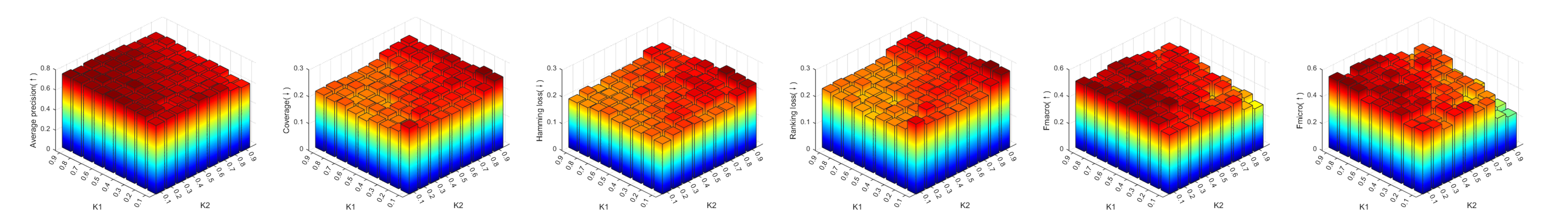}}
        \hfil
        \subfloat[\footnotesize Parameter sensitivity analysis of FedCMFS on the Image dataset when the number of clients is 10.]{\includegraphics[width=0.8\linewidth]{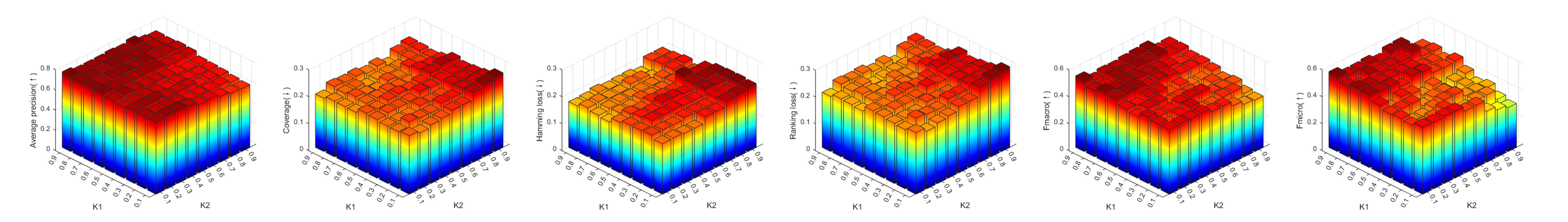}}
        \caption{Parameter sensitivity analysis of FedCMFS. }
        \label{Sensitivity}
    \end{figure*}

    This section focuses on investigating the impact of parameters $k_1$ and $k_2$ on the FedCMFS using two datasets, Flags and Image. The analysis include diverse client numbers, with detailed outcomes presented in Figure~\ref{Sensitivity}.

    In the case of the low-dimensional Flags dataset, it was found that the parameter $k_1$ has a relatively small effect on the results, while $k_2$ has a negligible effect on the results. Specifically, when the number of clients is 3 and 5, setting $k_1$ in the range $\left[0.1,0.4\right]$ can obtain better experimental results. In contrast, for 10 customers, the optimal range of $k_1\in\left[0.4,0.9\right]$. This variation in results is attributed to the limited number of features in the Flags dataset, resulting in fewer feature bases filtered out by FedCFR and FedCFC, thus reducing the impact of $k_1$ and $k_2$. Therefore, it may be more effective to use only FedCFL and FedCFR for feature selection. In addition, the experimental results are affected by the number of clients, with an inversion of the effective parameter range at 10 clients. This variation may be caused by the uneven distribution of samples in the smaller dataset when simulating the federated environment.
   
    In the case of the high-dimensional Image dataset, the parameters  $k_1$ and $k_2$ have a significant impact on the experimental results. In all three client scenarios, the results improve significantly when $k_1$ is set within $[0.6,0.9]$ and $k_2$ is set within  $[0.1,0.3]$. This sensitivity is attributed to the large number of features of the dataset, which allows FedCFR and FedCFC to effectively process and screen more features due to the complex associations present in the high-dimensional data. By correcting the top 10\% to 30\%  of possible erroneous features, FedCFR achieves the best performance. However, increasing the number of corrected features may lead to counterproductive adjustments, caused by excessive noise and strict symmetry constraints. Moreover, the complexity of the image dataset means that the data distribution is unlikely to be affected by the sampling method used to simulate the federated environment, making the number of clients irrelevant to the experimental results. The consistency of parameter sensitivities across different clients further demonstrates the effectiveness of FedCFR and FedCFC in the federated context.

\subsection{Statistical Hypothesis Testing}\label{sec5_7}
    To fully establish the superiority of FedCMFS over prior methods, we conduct the Friedman test ($\alpha = 0.05$) on the six metrics~\cite{demvsar2006statistical}. Table~\ref{tab9} shows the specific results. We observe that the Friedman statistic values on all metrics are higher than the critical value, which means the null hypothesis of no significant difference among the algorithms is rejected.

    \begin{table}[htbp]
        \caption{Friedman test results.\label{tab9}}
        \centering
        \begin{tabular}{ccc}
        \hline 
        \textbf{Metric} & $\mathbf{F_F}$  & \textbf{Critical Value($\alpha = 0.05$)}\\
        \hline 
        AvP         & 18.2674       & \multirow{6}{*}{2.293}  \\
        Cov         & 25.0398       & \\
        HaL         & 9.4003        & \\
        RaL         & 17.9430       & \\
        Macro-F1    & 17.4678       & \\
        Micro-F1    & 15.0658       & \\
        \hline 
        \end{tabular}
    \end{table}

\begin{figure*}[htb]
        \centering
        \subfloat[\footnotesize Average Precision]{\includegraphics[width=\textwidth/4]{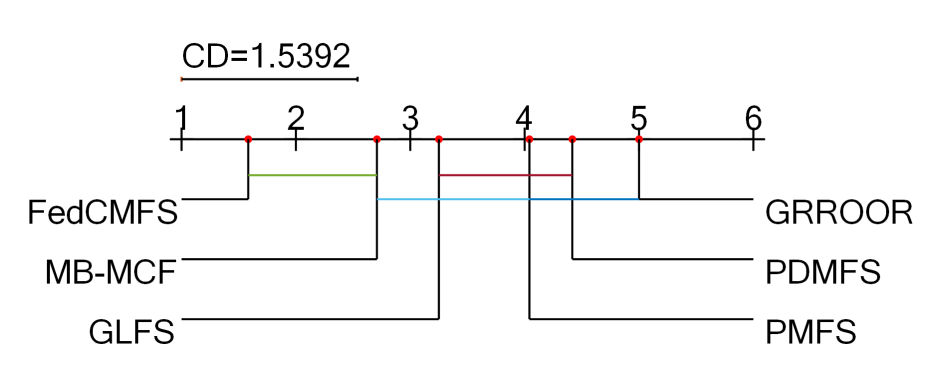}}
        \hfil
        \subfloat[\footnotesize Coverage]{\includegraphics[width=\textwidth/4]{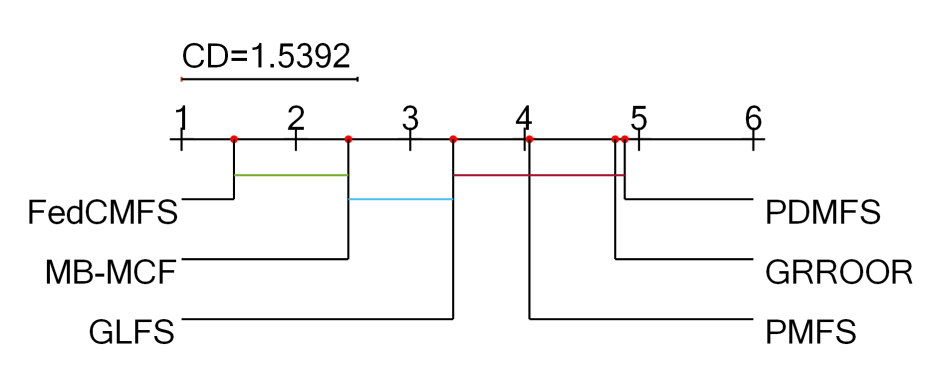}}
        \hfil
        \subfloat[\footnotesize Hamming Loss]{\includegraphics[width=\textwidth/4]{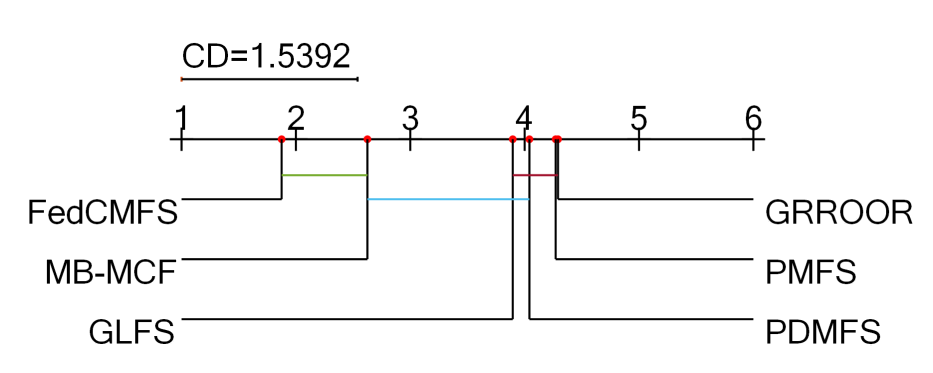}}
        \hfil
        \subfloat[\footnotesize Ranking Loss]{\includegraphics[width=\textwidth/4]{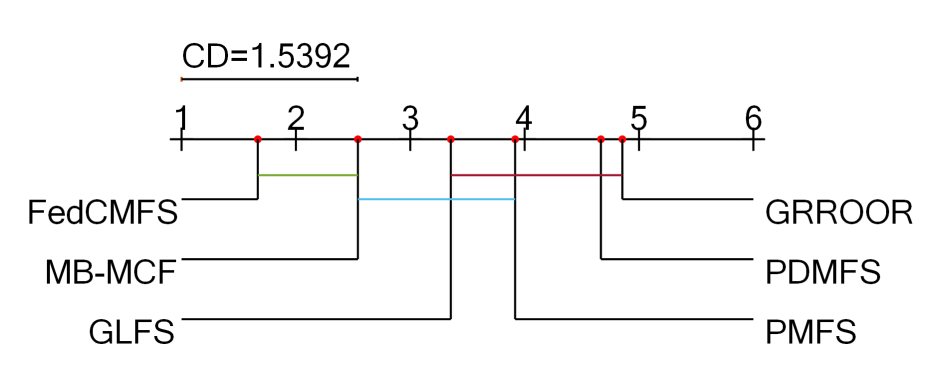}}
        \hfil
        \subfloat[\footnotesize Macro-F1]{\includegraphics[width=\textwidth/4]{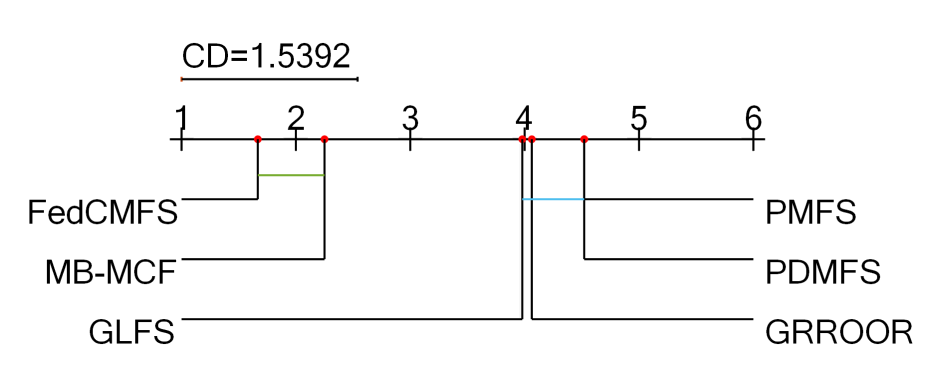}}
        \hfil
        \subfloat[\footnotesize Micro-F1]{\includegraphics[width=\textwidth/4]{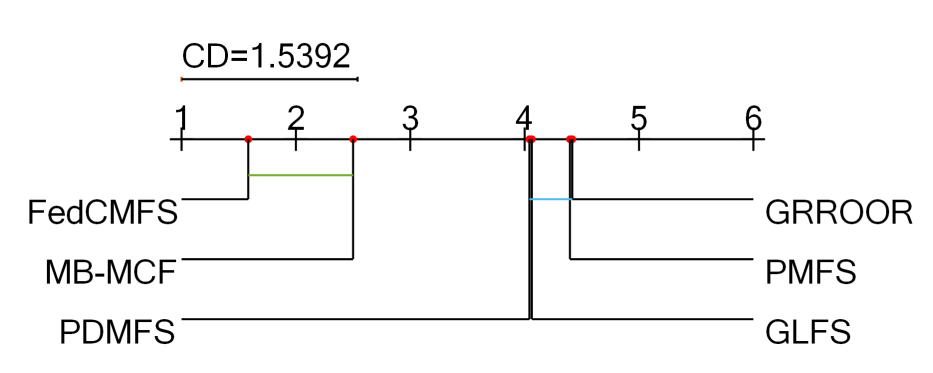}}
        \caption{Nemenyi Test Comparison between FedCMFS and Competitors.}
        \label{Statistical_Hypothesis_Testing}
    \end{figure*}

    Since rejecting the null hypothesis, we further employs the Nemenyi test~\cite{demvsar2006statistical} as a post-hoc test. The Nemenyi test indicates a significant difference in the performance of two methods if the mean rank difference between them exceed a critical difference (CD). The results are shown in Figure~\ref{Statistical_Hypothesis_Testing}, where each rank is sequentially marked on the axis and the lowest is on the right. Notably, FedCMFS achieves the lowest rank across all metrics and significantly outperforms other methods.

\subsection{Acceleration Effect}\label{sec5_8}
    To evaluate the effectiveness of the acceleration methods, we compared the runtime before and after applying the acceleration methods. As shown in Table \ref{tab10}, the acceleration effect is particularly noticeable for high-dimensional datasets such as Slashdot and Business. However, for the low-dimensional dataset Flags, the runtime after acceleration is longer than before. This anomaly is due to the relatively high overhead associated with GPU initialization and communication, which outweighs the shorter computation time. Next, we conducted a comparative analysis of its runtime against that of other algorithms. The results are summarized in the table~\ref{tab11}. After acceleration, the runtime of FedCMFS is close to that of PDMFS, outperforming MB-MCF, and demonstrating excellent performance on high-dimensional datasets.
    
    \begin{table}[htbp]
        \centering
        \scriptsize
        \caption{Acceleration Results and Speedup.}
            \setlength{\tabcolsep}{3.0pt}
        \begin{tabular}{ccccc}
            \hline
            Datasets & Client number & Original FedCMFS(s) & Accelerated FedCMFS(s) & Speedup \\ 
            \hline
            \multirow{3}{*}{VirusGo}  & 3  & 35.0103 & 6.1075 & 5.7323 \\
            & 5  & 63.6809 & 9.5575 & 6.6630 \\
            & 10 & 114.7019 & 18.0596 & 6.3513 \\ 
            \hline
            \multirow{3}{*}{Flags}    & 3  & 2.6883 & 3.0977 & 0.8678 \\
            & 5  & 3.0073 & 3.5928 & 0.8370 \\
            & 10 & 8.0959 & 4.9475 & 1.6364 \\ 
            \hline
            \multirow{3}{*}{Image}    & 3  & 107.6387 & 10.9457 & 9.8339 \\
            & 5  & 296.3440 & 20.4531 & 14.4890 \\
            & 10 & 503.1186 & 27.3849 & 18.3721 \\ 
            \hline
            \multirow{3}{*}{Slashdot} & 3  & 5121.2536 & 62.9691 & 81.3296 \\
            & 5  & 8985.2121 & 103.4271 & 86.8748 \\
            & 10 & 18305.3374 & 195.4889 & 93.6388 \\ 
            \hline
            \multirow{3}{*}{Business} & 3  & 4323.8704 & 103.6646 & 41.7102 \\
            & 5  & 6527.8086 & 144.6297 & 45.1346\\
            & 10 & 16983.4118 & 337.8335 & 50.2715 \\ 
            \hline
        \end{tabular}
        \label{tab10}
    \end{table}
    
    \begin{table}[htbp]
        \centering
        \setlength{\tabcolsep}{3.1pt}
        \scriptsize 
        \caption{Experimental result of Time(s) (↓).}
        \begin{tabular}{cccccccc}
            \hline
            Datasets & Clients & FedCMFS & MB-MCF & GLFS & PDMFS & GRROOR & PMFS \\ 
            \hline
            \multirow{3}{*}{VirusGo}  & 3  & 6.1075 & 7.2146 & 2.6246 & 42.1955 & 17.7764 & 1.1384 \\
            & 5  & 9.5575 & 10.8515 & 2.3958 & 79.6399 & 29.3349 & 0.6189 \\
            & 10 & 18.0596 & 19.6982 & 5.1917 & 149.7295 & 63.3173 & 1.3123 \\ 
            \hline
            \multirow{3}{*}{Flags}    & 3  & 3.0977 & 0.3931 & 0.1060 & 0.2187 & 0.2147 & 0.0837 \\
            & 5  & 3.5928 & 0.5043 & 0.1291 & 0.2988 & 0.5543 & 0.0270 \\
            & 10 & 4.9475 & 1.3031 & 0.3207 & 0.7390 & 0.8962 & 0.1349 \\ 
            \hline
            \multirow{3}{*}{Image}    & 3  & 10.9457 & 102.3162 & 2.2511 & 44.1345 & 4.0944 & 0.1340 \\
            & 5  & 20.4531 & 221.9948 & 4.7947 & 86.7157 & 8.0327 & 0.2245 \\
            & 10 & 27.3849 & 370.4468 & 8.7092 & 166.2046 & 14.4684 & 0.4853 \\ 
            \hline
            \multirow{3}{*}{Slashdot} & 3  & 62.9691 & 1872.0302 & 9.3750 & 597.7361 & 66.8130 &  1.5003\\
            & 5  & 103.4271 & 2579.8462 & 20.7131 & 900.6759 & 104.1713 & 2.3110 \\
            & 10 & 195.4889 & 5953.3369 & 28.4496 & 1842.9548 & 228.8878 & 4.9612 \\ 
            \hline
            \multirow{3}{*}{Business} & 3  & 103.6646 & 945.8063 & 2.3636 & 117.6500 & 10.2650 & 0.2526 \\
            & 5  & 144.6297 & 1504.0571 & 2.8750 & 191.7148 & 17.1086 & 0.4465 \\
            & 10 & 337.8335 & 3096.6454 & 5.5434 & 408.8154 & 36.7804 & 0.9499 \\ 
            \hline
        \end{tabular}
        \label{tab11}
    \end{table}
\section{Conclusion and Further Work}\label{sec6}
    To solve the problem of causal multi-label feature selection in  federated setting, this paper proposes the FedCMFS algorithm based on the local causal structure learning method and horizontal federated learning framework. 

 The experimental results show that FedCMFS achieves the best experimental results in federated setting. Specifically, FedCMFS is able to directly determine the parent-child relationship of variables by mining the causal relationship between variables, thus providing excellent interpretability. Second, the FedCMFS algorithm operates in a distributed data environment and maintains data privacy during the transmission of encrypted semantics. Finally, FedCMFS is also able to effectively correct the effects of noise and differences in client data quality on the algorithm performance. However, there are still issues that need to be further investigated to advance the field. We observe that FedCMFS, which is based on statistical methods, may incorrectly learn the causal relationships when a dataset has small-sized data samples, leading to incorrectly selected features. Therefore, exploring federated causal multi-label feature selection in small-sized datasets is also a promising research direction~\cite{xiang2023bootstrap}.

\bibliographystyle{IEEEtran}
\bibliography{reference}
\end{document}